%% file: root.tex
\documentclass[journal]{IEEEtran}

\IEEEoverridecommandlockouts                              

\usepackage{booktabs}

\usepackage[noadjust]{cite}
\nocite{*}  

\makeatletter
\let\NAT@parse\undefined
\makeatother

\usepackage{hyperref}
\usepackage{subcaption}
\usepackage{pdfpages}
\usepackage{balance}
\usepackage{tabularx}
\usepackage{adjustbox}
\usepackage{comment}
\hypersetup{
    colorlinks=true,
    citecolor=[RGB]{0,0,0},
    linkcolor=[RGB]{0,0,0},
    filecolor=[RGB]{0,0,0},      
    urlcolor=[RGB]{0,0,0},
    pdfpagemode=FullScreen,
}
\usepackage{placeins} 

\input{common}

\graphicspath{{figures/}}

\title{
\LARGE \bf
RCG: Safety-Critical Scenario Generation for Robust Autonomous Driving via Real-World Crash Grounding
}

\author{Benjamin Stoler, Juliet Yang, Jonathan Francis, and Jean Oh
\thanks{BS, JY, JF, and JO are with the School of Computer Science at Carnegie Mellon University. Emails: {\tt\footnotesize \{bstoler, jzyang, jmf1, jeanoh\}@cs.cmu.edu}. JF is also with the Bosch Center for AI.}
\thanks{Manuscript received May xx, 2025.}
}

\begin{document}

\markboth{Journal of \LaTeX\ Class Files,~Vol.~25, No.~7, July~2025}%
{Shell \MakeLowercase{\textit{et al.}}: Bare Demo of IEEEtran.cls for IEEE Journals}
%

\maketitle

\showauthorcommenttrue

\renewcommand{\thefootnote}{\fnsymbol{footnote}}

\begin{abstract}

Safety-critical scenarios are essential for training and evaluating autonomous driving (AD) systems, yet remain extremely rare in real-world driving datasets. To address this, we propose Real-world Crash Grounding (RCG), a scenario generation framework that integrates crash-informed semantics into adversarial perturbation pipelines. We construct a safety-aware behavior representation through contrastive pre-training on large-scale driving logs, followed by fine-tuning on a small, crash-rich dataset with approximate trajectory annotations extracted from video. This embedding captures semantic structure aligned with real-world accident behaviors and supports selection of adversary trajectories that are both high-risk and behaviorally realistic. We incorporate the resulting selection mechanism into two prior scenario generation pipelines, replacing their handcrafted scoring objectives with an embedding-based criterion. Experimental results show that ego agents trained against these generated scenarios achieve consistently higher downstream success rates, with an average improvement of 9.2\% across seven evaluation settings. Qualitative and quantitative analyses further demonstrate that our approach produces more plausible and nuanced adversary behaviors, enabling more effective and realistic stress testing of AD systems. Code and tools will be released publicly.

\end{abstract}


\input{sections/1_introduction}

\input{sections/2_related_works}

\input{sections/3_preliminaries}

\input{sections/4_dataset}

\input{sections/5_approach}

\input{sections/6_advgen}

\input{sections/7_experimental_setup}

\input{sections/8_results}

\input{sections/9_conclusion}


\addtolength{\textheight}{0cm}   








\bibliographystyle{IEEEtran}
\bstctlcite{IEEEexample:BSTcontrol}
\bibliography{ref}

\begin{IEEEbiographynophoto}{Benjamin Stoler} is a Ph.D. candidate in the Computer Science Department at Carnegie Mellon University. Benjamin's research focuses on safety and robustness in autonomous driving and social robot navigation. Benjamin obtained his B.S. and M.S. degrees in Computer Science from the College of Engineering at the University of Michigan.
\end{IEEEbiographynophoto}

\begin{IEEEbiographynophoto}{Juliet Yang}
is an undergraduate student in the Computer Science Department at Carnegie Mellon University. Juliet is interested in exploring ways to improve the robustness of autonomous driving systems.
\end{IEEEbiographynophoto}

\begin{IEEEbiographynophoto}{Jonathan Francis} is a Senior Research Scientist at the Bosch Center for Artificial Intelligence (BCAI), Courtesy Faculty in the Robotics Institute at Carnegie Mellon University (CMU), and the Robotics Research Program Lead in the Carnegie-Bosch Institute, College of Engineering, at CMU. Jonathan's research focuses on developing autonomous systems that learn safe, robust, multimodal, and transferrable representations of the world and are therefore capable of acquiring, adapting, and improving their own skills/behavior when deployed to diverse scenarios. Jonathan received his Ph.D. from the School of Computer Science at CMU and his B.S. and M.S. in Electrical \& Computer Engineering at CMU.
\end{IEEEbiographynophoto}

\begin{IEEEbiographynophoto}{Jean Oh} is an Associate Research Professor at the Robotics Institute at Carnegie Mellon University (CMU) and head of the roBot Intelligence Group (BIG). Jean’s research goal is to develop technologies to remind us of “what makes us humans,” promoting humanity such as \textit{safety}, \textit{creativity}, and \textit{compassion} in various problem domains including self-driving, safe aviation, and arts. Jean’s works on social navigation and creative robotics have won several best paper awards at various robotics conferences such as IEEE International Conference on Robotics and Automation (ICRA) and ACM/IEEE International Conference on Human-Robot Interaction (HRI), and featured in media around the world including New York Times and CNET. Jean received PhD at CMU, MS at Columbia University, and BS at Yonsei University.
\end{IEEEbiographynophoto}

\end{document}

%% file: common.tex
\usepackage{graphicx}
\usepackage{hyperref}
\usepackage{keyval}
\usepackage{algorithm}
\usepackage{algpseudocode}
\usepackage{amsmath, amsfonts}
\usepackage{amssymb}
\usepackage{cleveref}
\usepackage{fancyhdr} 
\usepackage{color}
\usepackage{transparent}
\usepackage{booktabs}
\usepackage{tabularx}
\usepackage{xspace}

\usepackage{caption}
\usepackage{subcaption}
\usepackage{textcomp}
\usepackage{stfloats}
\usepackage{url}
\usepackage{verbatim}
\usepackage{graphicx}
\usepackage{multirow}
\usepackage{mathtools}

\usepackage{enumitem} 
\usepackage{booktabs}

\usepackage{caption}
\usepackage{tabularx}
\usepackage{threeparttable}
\usepackage{makecell}

\newcommand{\idest}{i.e.,\xspace}
\newcommand{\exempli}{e.g.,\xspace}

\newcommand{\ego}{\texttt{ego}\xspace}
\newcommand{\adv}{\texttt{adv}\xspace}

\newif\ifshowstatshorizontal
\showstatshorizontaltrue

\newif\ifshowauthorcomment
\showauthorcommenttrue

\newcommand{\para}[1]{\noindent\textbf{#1}}

\usepackage{xargs} 
\usepackage{xcolor}
\definecolor{safegreen}{RGB}{0,100,0}
\definecolor{neutralblue}{RGB}{0,0,139}
\definecolor{unsafeorange}{RGB}{255,140,0}
\definecolor{lilac}{RGB}{180,145,215}

\usepackage[colorinlistoftodos,prependcaption,textsize=tiny]{todonotes}
\newcommandx{\unsure}[2][1=]{\todo[linecolor=red,backgroundcolor=red!25,bordercolor=red,#1]{#2}}
\newcommandx{\change}[2][1=]{\todo[linecolor=blue,backgroundcolor=blue!25,bordercolor=blue,#1]{#2}}
\newcommandx{\info}[2][1=]{\todo[linecolor=OliveGreen,backgroundcolor=OliveGreen!25,bordercolor=OliveGreen,#1]{#2}}
\newcommandx{\improvement}[2][1=]{\todo[linecolor=Plum,backgroundcolor=Plum!25,bordercolor=Plum,#1]{#2}}
\newcommandx{\thiswillnotshow}[2][1=]{\todo[disable,#1]{#2}}
 
\def\HiLiYellow{\leavevmode\rlap{\hbox to \hsize{\color{yellow!20}\leaders\hrule height .8\baselineskip depth .4ex\hfill}}}
  
\fancypagestyle{firststyle}
{
  \fancyhf{}

}

\fancypagestyle{secondstyle}
{
  \fancyhf{}
  \fancyhead[R]{\footnotesize \thepage}

}

\usepackage{cleveref}

%% file: sections/1_introduction.tex
\section{Introduction} \label{sec:introduction}

\IEEEPARstart{T}{o} develop and assess autonomous driving (AD) models, a diverse and representative set of traffic scenarios must be used to ensure that such agents behave safely in real-world cases, especially in long-tail, safety-critical scenarios~\cite{hauer2019did, stoler2024safeshift, sun2021scenario}. To reduce reliance on costly and potentially risky real-world driving testing, researchers and industry practitioners lean on large datasets of collected driving logs~\cite{webb2020waymo, kalra2016driving}. These datasets typically provide agent-centric, trajectory-level motion annotations, enabling supervised learning of motion forecasting and simulation-based driving tasks~\cite{ettinger2021large, caesar2021nuplan, wilson2023argoverse}. However, safety-critical scenarios, such as actual accident and near-miss cases, are exceedingly scarce in these standard motion datasets~\cite{liu2024curse, huang2025cadre, ding2023survey}. To address this phenomenon known as ``the curse of rarity,'' much recent work has turned to adversarial scenario generation to produce these rare events synthetically~\cite{zhang2023cat,ransiek2024goose,stoler2024seal}. Modifying (or ``perturbing'') these benign scenarios is appealing because it retains realistic context from real-world scenes while allowing adversaries to be optimized to trigger failure in fine-grained ways.

To adversarially perturb a scenario with respect to a driving model under development or test (\idest the ``ego'' agent controlled by that model), it would be ideal to be able to \textit{directly} sample appropriate, context-dependent adversary behaviors that sufficiently challenge the ego agent's decision making.
However, due to the curse of rarity, prior art must instead cope in alternative ways. In particular, recent state-of-the-art (SOTA) works such as CAT~\cite{zhang2023cat} and SEAL~\cite{stoler2024seal} \textit{reason} over candidate adversary behaviors, produced by a pre-trained trajectory generator, with respect to expert criteria (\exempli attempting to maximize collision closeness to the ego, or forcing the ego to perform harsh avoidance maneuvers, etc.). Despite these advances, generated scenarios by such approaches tend to still be overly aggressive and thus too far from being realistic.

To generate realistic safety-critical scenarios, we propose \textbf{R}eal-world \textbf{C}rash \textbf{G}rounding (RCG), which begins by leveraging a large, well-annotated dataset of everyday driving to model moderately unsafe behaviors. To capture more extreme but realistic failures, we reach beyond conventional datasets for autonomous driving prediction and control and we draw from minimally-annotated video corpora originally collected for traffic scene understanding tasks, which include real crash and near-miss cases~\cite{chai2024tads, xu2021sutd}. This raises the key question of how to meaningfully incorporate such non-aligned data; RCG addresses this by using a representation learning framework to unify and structure behaviors across both sources.

This representation is constructed in two stages. First, we collect and heuristically classify large-scale ``safe'', ``neutral'', and ``unsafe'' behaviors, from everyday but safety-relevant scenarios. We train the embedding using a prototypical contrastive learning (PCL)~\cite{li2021prototypical} objective that encourages local structure between contextually similar behaviors while separating distinct safety classes. This enables pre-training over diverse scene contexts while defining a structured  notion of ``unsafe'' behavior. Second, we fine-tune the representation on a small number of real accident and near-miss interactions. Since these examples lack motion annotations, we use off-the-shelf perception tools to extract approximate trajectories that meaningfully refine the embedding around real-world unsafe behaviors. 

Once trained, our embedding space yields an adversarial selection process by quantifying how closely a generated trajectory matches real unsafe behaviors in context, ultimately serving as an optimization objective. This objective can then be easily integrated on top of existing scenario generation approaches. By guiding adversarial agents to maximize realistic criticality, we generate safety-critical scenarios that are both more plausible as well as effective for training and evaluating driving models. Thus, RCG also addresses a broader methodological challenge: using representation learning to extract useful structure from non-aligned, weakly-labeled data, enabling it to inform downstream tasks despite minimal task-specific annotation.

\textbf{Our contributions} are thus as follows:
\begin{enumerate}
    \item A safety-informed representation space that captures contextualized driving behaviors with global and local structure, and enables integration of non-aligned data sources not previously used or intended for this task.
    \item A crash-grounded scoring objective for adversarial scenario generators that improves the plausibility and practical usability of generated scenarios. 
    \item Extensive validation showing that ego agents trained in closed-loop with our generated adversaries display more robust performance, improving driving success rates by an average of $9.2\%$ across all tested environments and base scenario generation approaches.
\end{enumerate}

%% file: sections/2_related_works.tex
\section{Related Work} \label{sec:related_works}

\subsection{Autonomous Driving Scenario Curation}

Much work in autonomous driving has focused on curating large-scale real-world driving logs for training and evaluating perception, prediction, and planning modules~\cite{wilson2023argoverse, ettinger2021large, caesar2021nuplan}. These datasets are typically multi-modal, with sensor data and annotations tailored to specific downstream tasks. For instance, perception often relies on image and LiDAR inputs with dense labels~\cite{sun2020scalability, zhang2022openmpd, caesar2020nuscenes}, while prediction and planning tasks utilize agent trajectories and map features, often in polyline form~\cite{krajewski2018highd, wilson2023argoverse, ettinger2021large}. While rich and widely used, such datasets overwhelmingly capture nominal behavior and suffer from the well-known ``curse of rarity,'' \idest the near-absence of safety-critical events~\cite{liu2024curse, ding2023survey, huang2025cadre}. Efforts like SafeShift~\cite{stoler2024safeshift} attempt to surface more safety-relevant scenes from these corpora, but truly critical interactions remain sparse or entirely missing.

To address this gap, several datasets have been released that capture real-world accidents or near-misses, often from dashcams or fixed-position surveillance footage. Ego-centric video datasets such as MM-AU~\cite{fang2024abductive} focus on reasoning and forecasting tasks, while others like TADS~\cite{chai2024tads}, SUTD~\cite{xu2021sutd}, and CADP~\cite{shah2018cadp} compile surveillance footage of crashes with varying levels of annotation. These datasets are typically small in scale and offer coarse labels like crash timing, agent roles; only a few, \exempli ~\cite{yan2023learning}, include a small number of trajectory-level annotations, which are required for most driving behavior learning approaches. In this work, we approximately annotate safety-critical examples from TADS~\cite{chai2024tads} and incorporate them as a fine-tuning stage in a contrastive learning pipeline, enabling semantic grounding on real-world accident structure without requiring full supervision.

\subsection{Closed-Loop and Adversarial Scenario Generation}

Alongside dataset curation, much work has explored generating synthetic autonomous driving scenarios directly. While many methods focus on reproducing normal driving behavior~\cite{suo2023mixsim, xu2023bits, ge2024task}, generating adversarial behavior is far more challenging due to the curse of rarity and distribution shift to benign scenarios, and must carefully balance diversity, fidelity, and usability for ego training~\cite{stoler2024seal, ding2023survey}. Some approaches manually construct starting scenarios based on real-world accident patterns~\cite{scanlon2021waymo, zhu2021hazardous, li2023advanced, ding2024realgen}, but these often require significant effort and domain expertise, and are limited to a narrow set of scenario types.

Thus, many approaches have turned to adversarial perturbation of real driving data, though these too face limitations. Several works rely on diffusion or black-box optimization methods that are computationally intensive and impractical for closed-loop ego policy development~\cite{xu2023diffscene, chang2024safesim, rempe2022generating, suo2023mixsim}. More efficient alternatives employ quality-diversity optimization~\cite{huang2025cadre} or adversarial reinforcement learning~\cite{ransiek2024goose}, but depend on hand-crafted metric functions that are difficult to generalize. Similarly, explicitly closed-loop methods such as CAT~\cite{zhang2023cat} and SEAL~\cite{stoler2024seal} select behaviors from lightweight generative priors, but over-optimize for collision proximity or induced ego deviation, often resulting in unrealistic adversary behavior. Our work follows this closed-loop adversarial selection paradigm, while emphasizing perturbations \textit{grounded} in a more generalized representation of real-world crash structure.

\subsection{Representation Learning for Driving Behavior}

Learning representations of agent behavior is a foundational component of many autonomous driving systems, particularly in tasks such as motion forecasting and simulation-based policy learning. A common approach leverages encoder-decoder architectures that \textit{implicitly} capture driving semantics in a latent bottleneck~\cite{casas2020implicit, zhang2023real, shi2022motion, wu2024smart}. These methods often aim to build expressive embedding spaces using high-capacity models such as Transformers~\cite{vaswani2017attention} or large-language-inspired tokenization schemes~\cite{tian2024tokenize}. Recent work has also explored the use of skill-space priors to enable more human-like, hierarchical behavior generation~\cite{hao2024skill, rana2023residual}. Other approaches attempt to embed safety awareness directly, either by incorporating risk-aware architectural components~\cite{wang2025risk} or by penalizing predicted collisions across trajectory samples~\cite{van2019safecritic}. While these techniques often improve downstream task performance, the resulting latent representations themselves remain difficult to interpret or align with safety-relevant structure.

To imbue learned embeddings with semantic structure more explicitly, contrastive learning has become an increasingly popular strategy in the machine learning community~\cite{hu2024comprehensive, gui2024survey}. In autonomous driving, recent works, such as FEND~\cite{wang2023fend} and TRACT~\cite{zhang2024tract}, apply prototypical contrastive learning (PCL)~\cite{li2021prototypical} to enforce structure based on trajectory types and training-stage dynamics. Other approaches, like LIDP~\cite{liu2024lidp}, rely on traditional contrastive objectives such as the InfoNCE loss~\cite{oord2018representation}, augmenting trajectory features to capture individualized driving patterns and style. Building on these ideas, we combine contrastive pre-training with crash-grounded fine-tuning to produce a representation space structured explicitly around safety-class separation, and further refined to reflect real-world accident behaviors.

%% file: sections/3_preliminaries.tex
\section{Preliminaries} \label{sec:preliminaries}

In closed-loop scenario generation, scenarios are created adaptively, based on how a learning agent interacts with an environment. We define notation here for the purposes of \textit{perturbation-based} safety-critical scenario generation, adapting definitions from SEAL~\cite{stoler2024seal}. In this setting, a base scenario from a trajectory-based, real-world dataset is selected and modified according to some desiderata pertaining to the learning agent, so as to increase the criticality of the ensuing interactions. The agent is then rolled out in the modified scenario, receiving some training impetus, and the cycle (\idest ``loop'') repeats. A base scenario is defined as $S=(\mathbf{X}, \mathbf{M}, \ego, \adv)$, where:

i) $\mathbf{X}=\{X_1, X_2, ..., X_N\}$ denotes the trajectories of all traffic participants (\idest ``agents'') in $S$. Each $X_i=\{x_i^{(t)}\}_{t=1}^T$ is the trajectory of agent $i$, consisting of its observed sequence of states over $T$ discrete timesteps, where $T$ is fixed and shared across agents within the scenario. Each state $x^{(t)}_i$ includes the agent's ground-plane position, heading, and velocity.

ii) $\mathbf{M}$ consists of the relevant \textit{map} context (\exempli road lines, stop sign locations, traffic light locations, etc.). This context includes primarily static information, as fixed locations in the ground-plane, but also includes dynamic elements such as the status of each traffic light at each timestep.

iii) \ego and \adv specify the agent IDs corresponding to the ego-controlled learning agent and the agent to be adversarially modified, respectively.

Now, let $\mathcal{P}$ denote a \emph{perturbation function}, which modifies a base scenario $S$ by altering the behavior of agent $\texttt{adv}$, possibly conditioned on historical cues. In a closed-loop training or evaluation cycle, the adversary is permitted to observe all trajectories from up to $K$ past roll-outs on the base scenario, denoted $\{\hat{\mathbf{X}}^{(k)}\}_{k=1}^{\leq K}$, each corresponding to a prior episode with a different perturbation of $S$. These roll-outs are themselves the result of rolling out the ego agent in response to prior adversarial perturbations, forming a sequential interaction history. Concretely, $\mathcal{P}$ is defined as follows: %
\begin{equation*}
\mathcal{P}: \left(S, \left\{\hat{\mathbf{X}}^{(k)}\right\}_{k=1}^{\leq K} \right) \rightarrow \mathcal{B}_{\texttt{adv}}
\end{equation*} %
\noindent where the planned adversarial behavior functional $\mathcal{B}_{\texttt{adv}}$ may take multiple forms, ranging from a fixed trajectory to follow, to a reactive policy capable of adapting to ego behavior while pursuing higher-level objectives.
Importantly, this perturbation process may condition on the original scenario, $S$, as well as past episode roll-outs, but must not access future planned ego trajectories for the upcoming episode.

%% file: sections/4_dataset.tex
\begin{figure*}[t]
\centering

\includegraphics[width=0.325\textwidth, trim=200 350 200 50, clip]{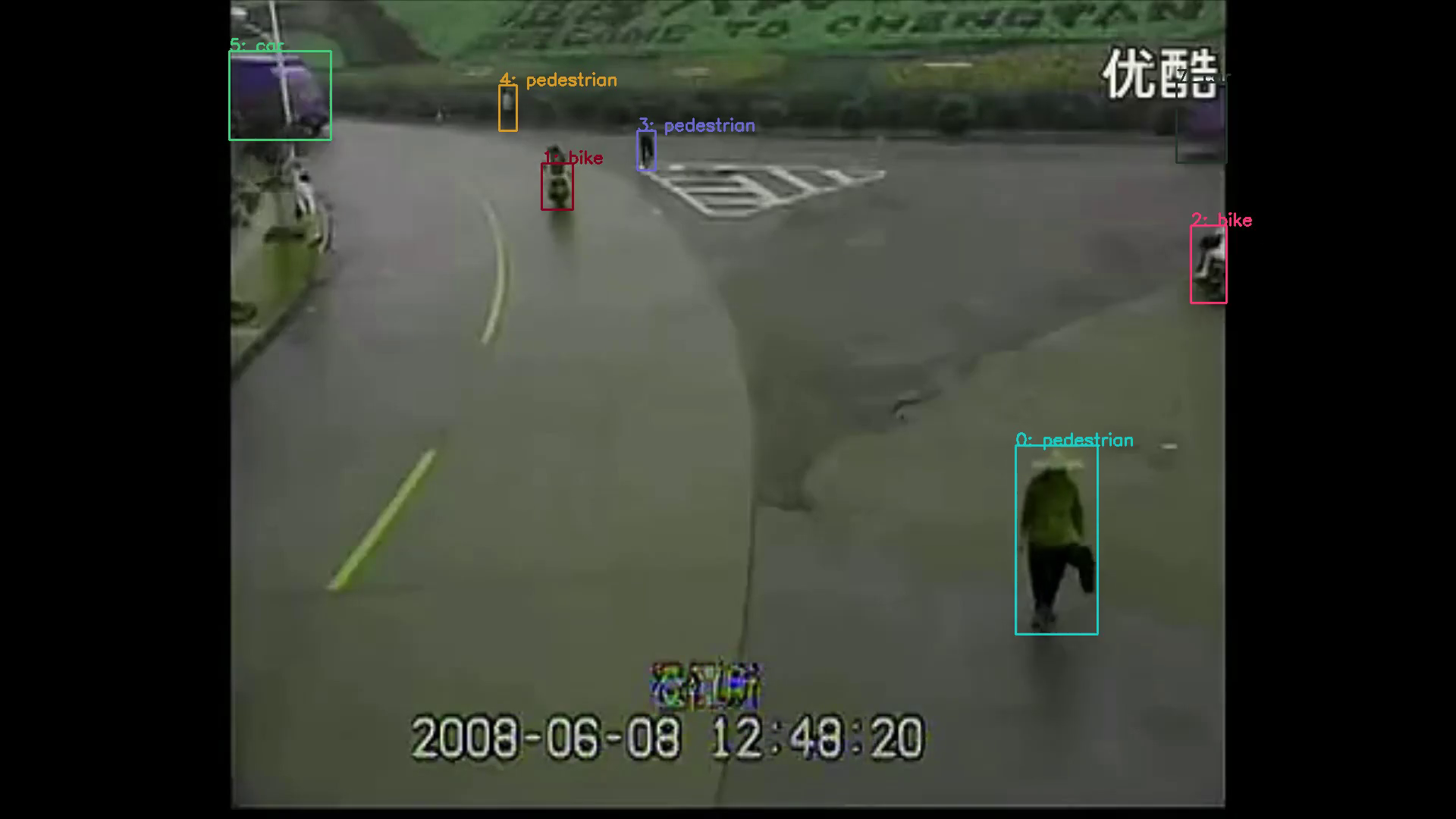}
\hfill
\includegraphics[width=0.325\textwidth, trim=200 350 200 50, clip]{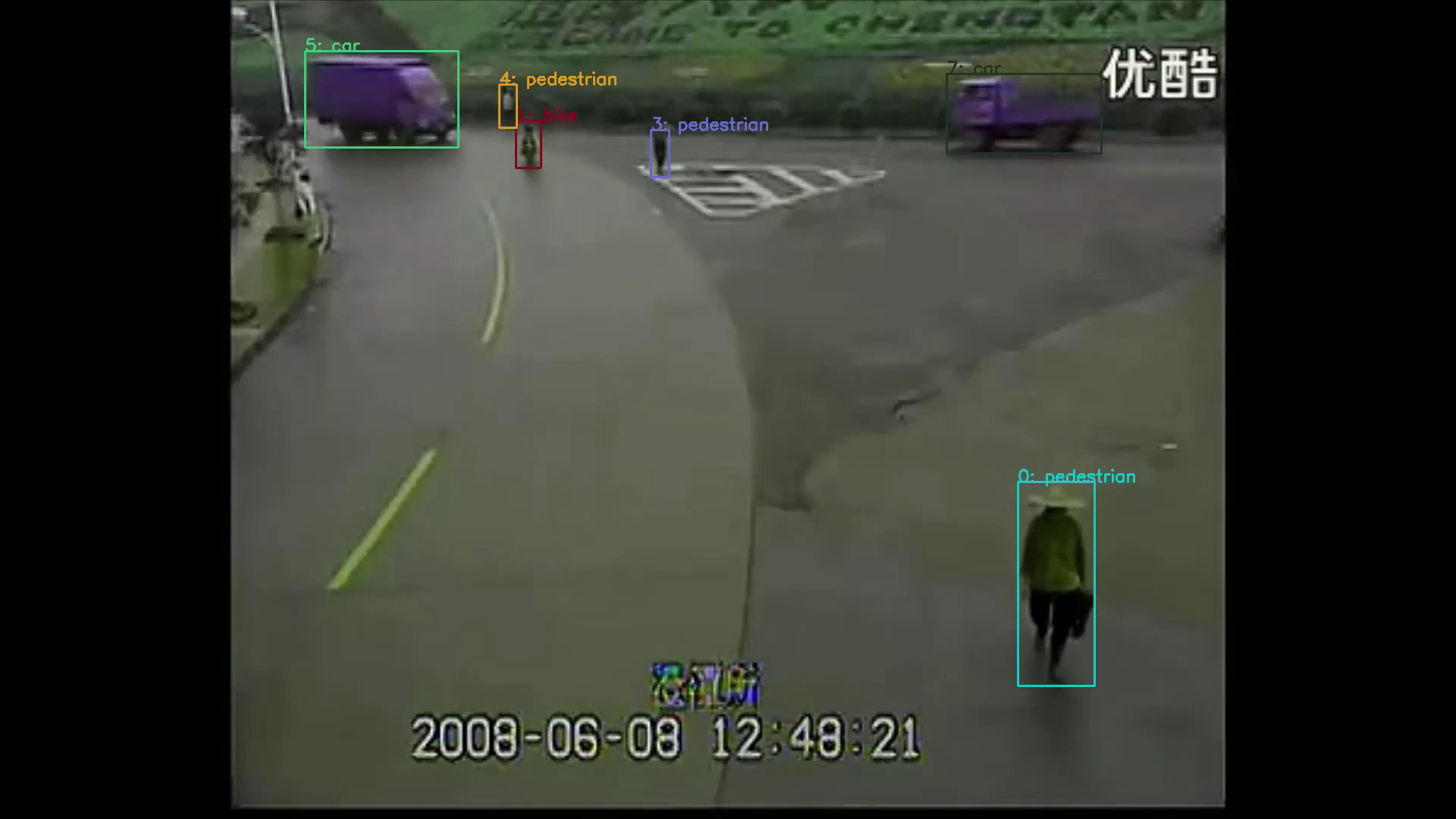}
\hfill
\includegraphics[width=0.325\textwidth, trim=200 350 200 50, clip]{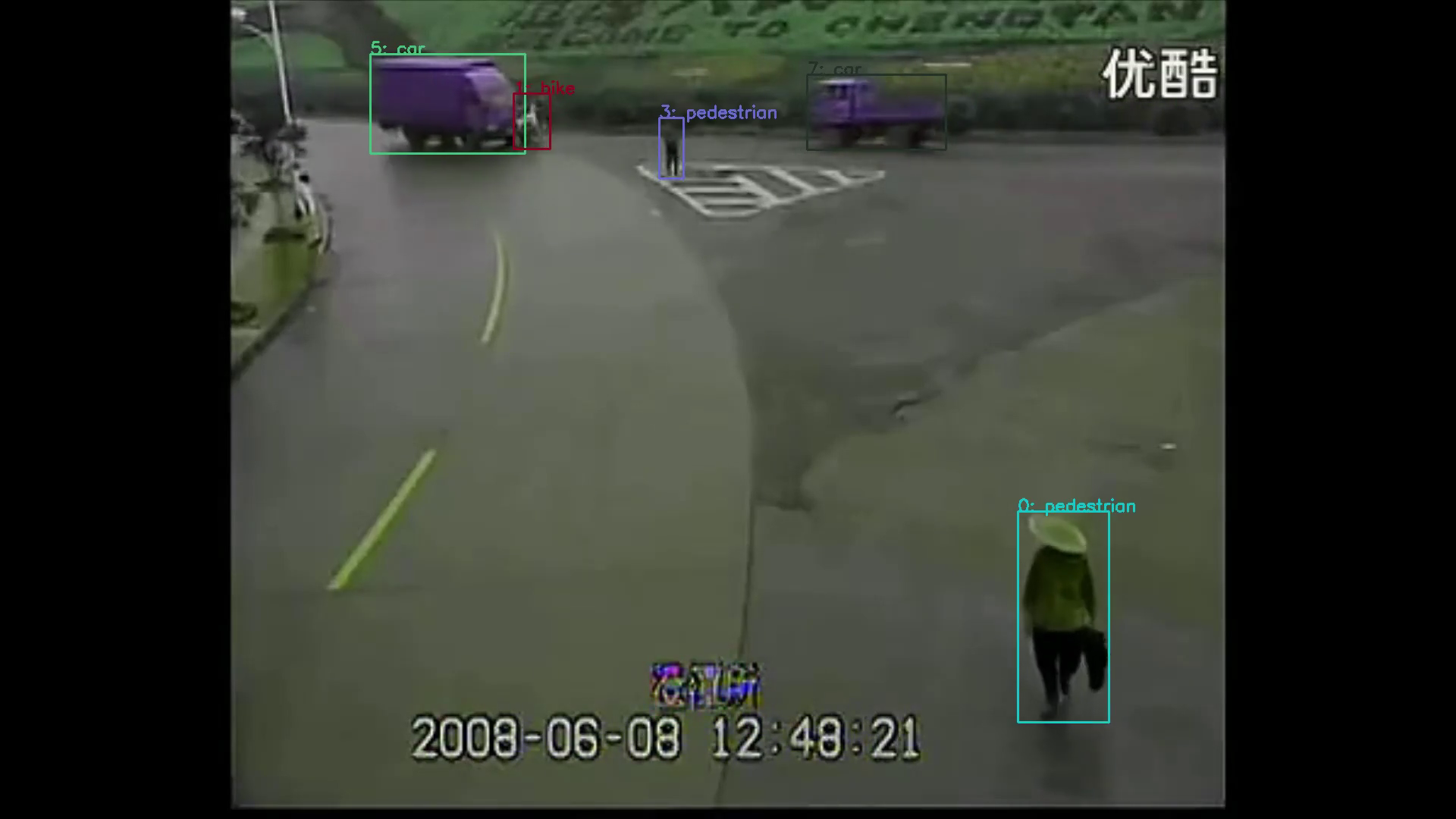}

\vspace{4mm}

\includegraphics[width=0.325\textwidth, trim=200 250 200 150, clip]{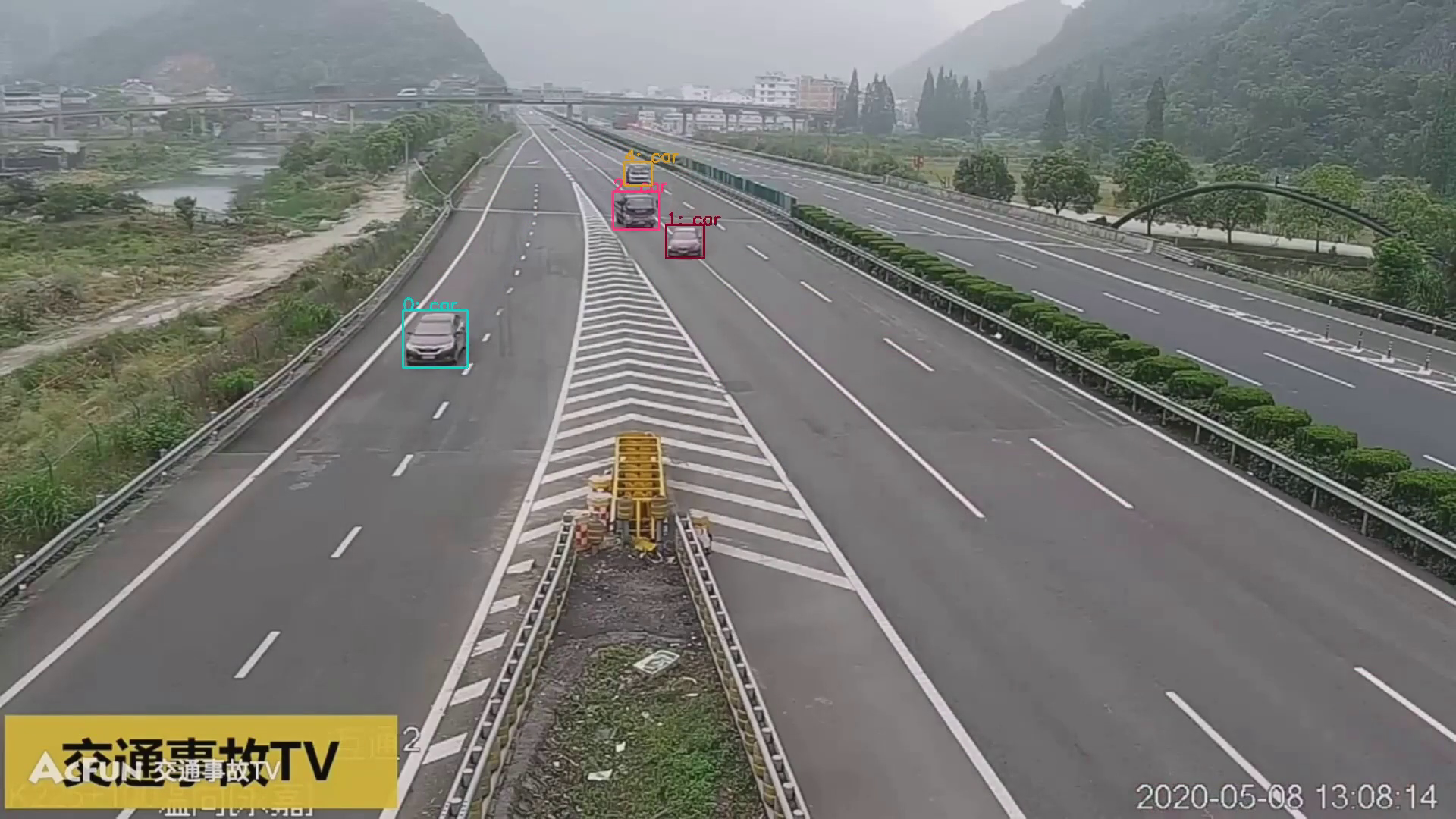}
\hfill
\includegraphics[width=0.325\textwidth, trim=200 250 200 150, clip]{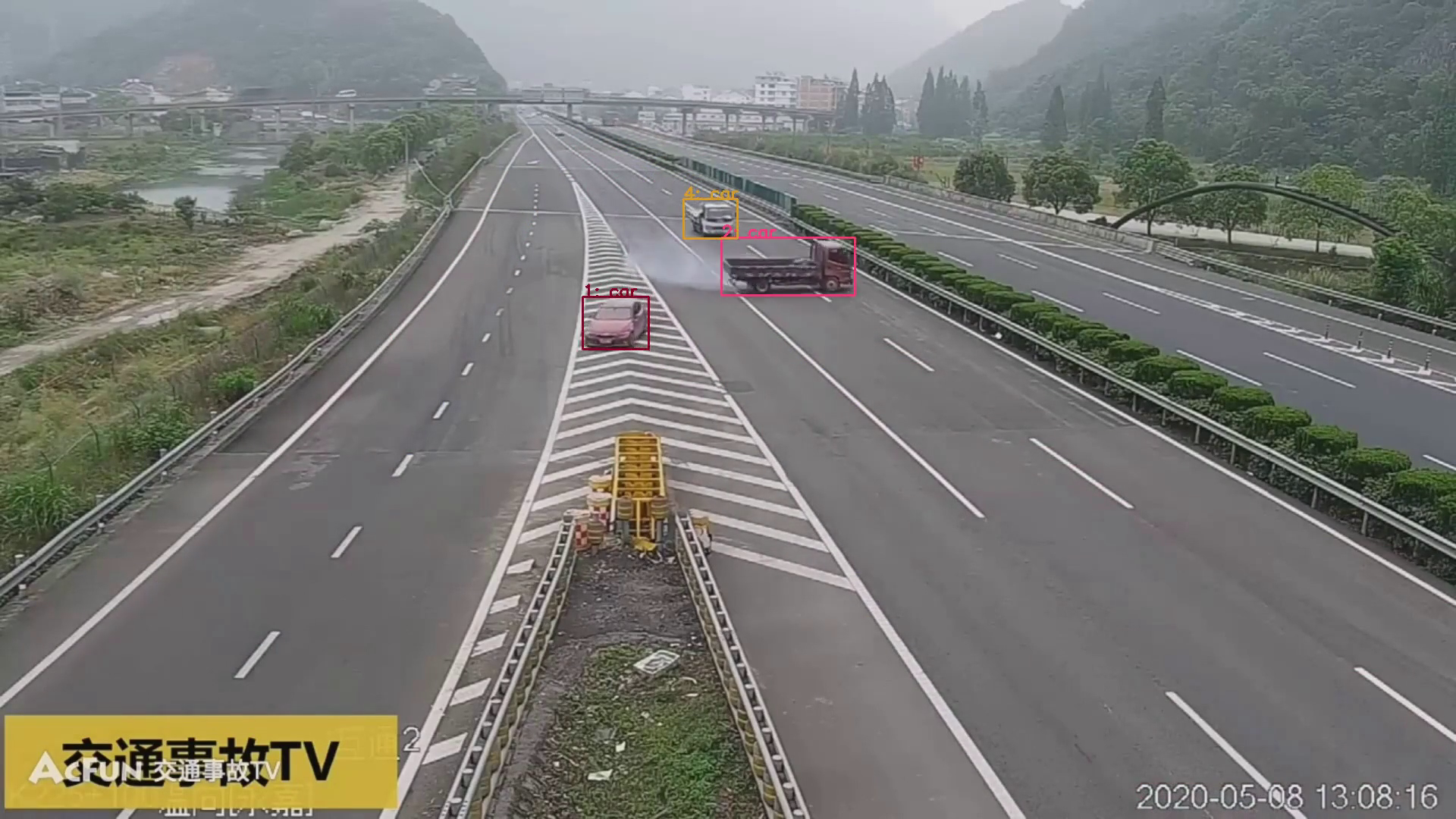}
\hfill
\includegraphics[width=0.325\textwidth, trim=200 250 200 150, clip]{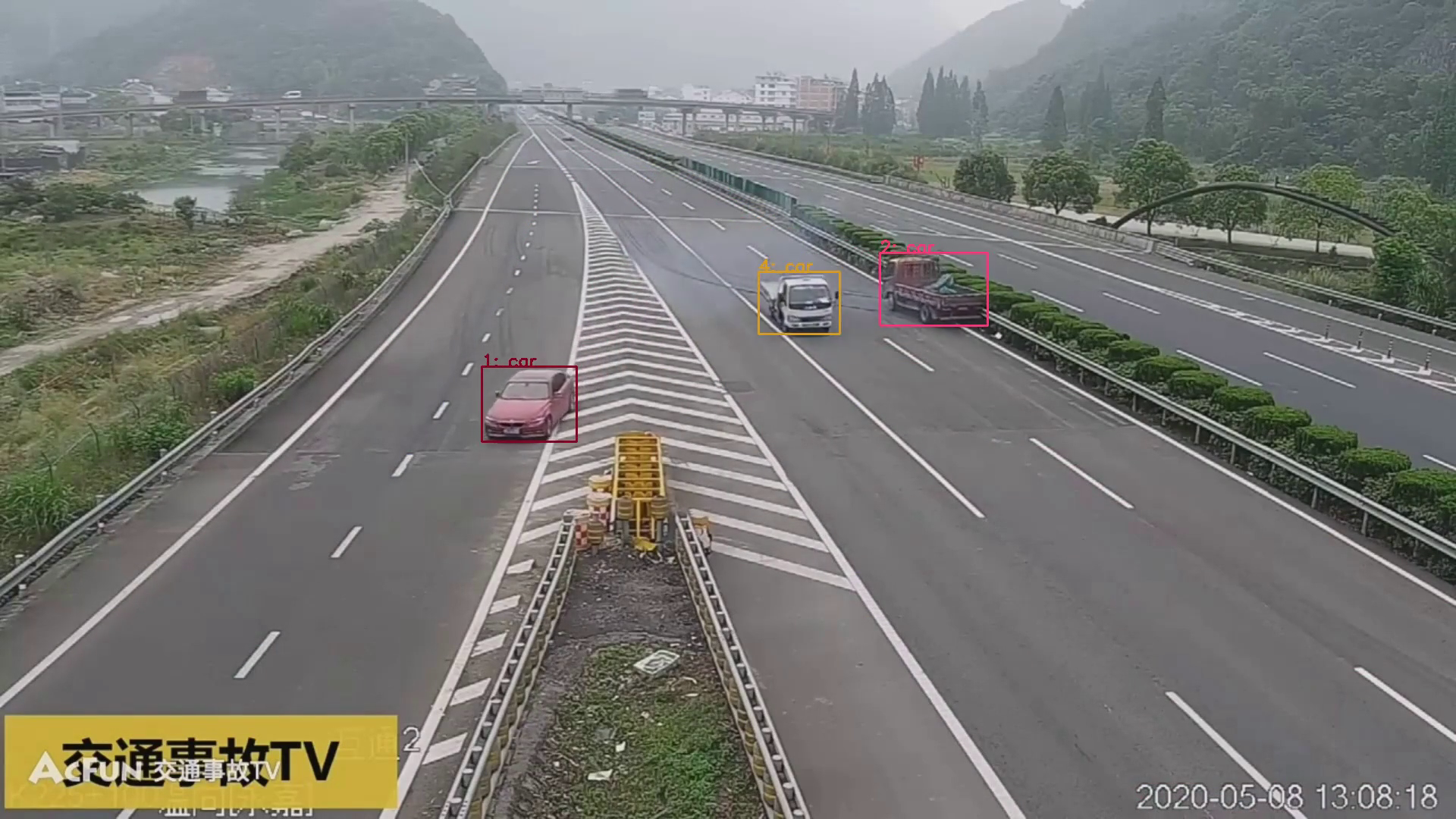}

\caption{Illustrative examples from TADS~\cite{chai2024tads}, processed into \texttt{TADS-traj}. In the top example, a vehicle (ID 5, ``safe'') can be seen approaching from the left, while executing a braking maneuver to reduce the severity of the collision, while a bike (ID 1, ``unsafe'') fails to react, increasing the criticality. In the bottom example, a vehicle (ID 2, ``unsafe'') swerves to avoid a car (ID 1, ``unsafe'') cutting in front of it, causing a third vehicle (ID 4, ``safe'') to swerve and brake to avoid an accident itself.} 
\label{fig:tads_examples}
\end{figure*}

\section{Accident Dataset Processing} \label{sec:dataset}

To provide safety-critical examples for constructing our embedding space, we curate a small set of real-world accident scenarios derived from the TADS dataset~\cite{chai2024tads}. Originally designed for road-traffic accident detection in CCTV footage, TADS comprises a relatively diverse collection of high-quality surveillance videos capturing vehicular accidents. These fixed, third-person perspectives enable more stable perception, which is essential to capturing the nuanced actions and reactions displayed by traffic participants in response to developing criticality. These types of interactions include subtle failures of anticipation, missed perceptions, and even actions taken by colliding participants to reduce the severity of an inevitable collision; crucially, such behaviors are not represented in large-scale trajectory-level datasets such as the Waymo Open Motion Dataset (WOMD)~\cite{ettinger2021large}, which are sourced from non-critical driving logs. Incorporating such accident scenarios is thus essential for modeling high-risk behaviors that lie outside the typical training distribution, and enabling grounded adversarial scenario-generation.

\para{Trajectory Extraction.} To extract trajectory-level annotations from TADS, we leverage recent advances in off-the-shelf perception tools, which make it increasingly feasible to process video-only datasets without extensive manual labeling. We apply a pipeline of recent foundation models, starting with GroundingDINO~\cite{liu2024grounding} for object detection and SAM2~\cite{ravi2024sam} for instance segmentation. Then, AED~\cite{fang2024associate} is used for 2D object tracking, while we utilize XFeat~\cite{potje2024xfeat} features to further refine temporal associations through keypoint correspondence. Next, metric depth is estimated with UniDepth~\cite{piccinelli2024unidepth} and temporally smoothed via VideoDepthAnything~\cite{chen2025video}, allowing us to follow OVM3D-Det~\cite{huang2024training} to estimate and orient 3D bounding boxes. We further smooth these agent bounding box sequences spatio-temporally to yield stable agent trajectories.

\para{Trajectory Filtering.} Following automated processing, we conduct a manual filtering and refinement step. From the roughly $1,000$ available videos in TADS, we exclude scenarios that are perception-degraded, feature unavoidable collisions, or contain insufficient context before the incident, selecting instead those that highlight subtle or reactive behaviors in the moments preceding an accident. This produces approximately 144 crash-containing \textit{scenarios}, from which we extract 385 accident-involved agent \textit{trajectories}. Each trajectory is manually labeled according to its role in the incident: ``aggressor", ``recipient", or ``bystander". In addition, we annotate one or more observed maneuvers per agent. Each maneuver is classified by both a discrete behavioral type (\idest brake, swerve, none, etc.) and a high-level safety class (safe, neutral, or unsafe), based on counterfactual reasoning about how the agent's active behavior impacted criticality.

A detailed breakdown of role and maneuver labels is shown in \Cref{tab:tads_stats}. We highlight representative examples of the scenarios in \Cref{fig:tads_examples}, illustrating the sort of complex, nuanced behaviors that are essential for training. We denote this processed and filtered version of TADS as \texttt{TADS-traj}.


\input{tables/tads_v1}

%% file: tables/tads_v1.tex
\begin{table}[t]
\centering
\caption{Distribution of annotated agent roles and observed maneuvers across 385 accident-involved agents. Note, multiple maneuvers may be assigned to a single agent.}
\resizebox{0.75\columnwidth}{!}{%
\begin{tabular}{llr}
\toprule
\textbf{Annotation Type} & \textbf{Category} & \textbf{Count} \\
\midrule
\multirow{3}{*}{Role} 
 & Aggressor & 155 \\
 & Recipient & 162 \\
 & Bystander & 68 \\
\midrule
\multirow{9}{*}{Maneuver} 
 & Neutral (none) & 40 \\
 & Neutral (brake) & 7 \\
 & Neutral (swerve) & 5 \\
 & Safe (brake) & 142 \\
 & Safe (swerve) & 86 \\
 & Safe (speed-up) & 5 \\
 & Safe (backup) & 1 \\
 & Unsafe (none) & 83 \\
 & Unsafe (brake) & 14 \\
 & Unsafe (swerve) & 20 \\
\bottomrule
\end{tabular}
}
\label{tab:tads_stats}
\end{table}

%% file: sections/5_approach.tex
\section{Learning A Safety-Informed Embedding Space} \label{sec:approach}

\begin{figure*}[t]
    \includegraphics[width=\textwidth]{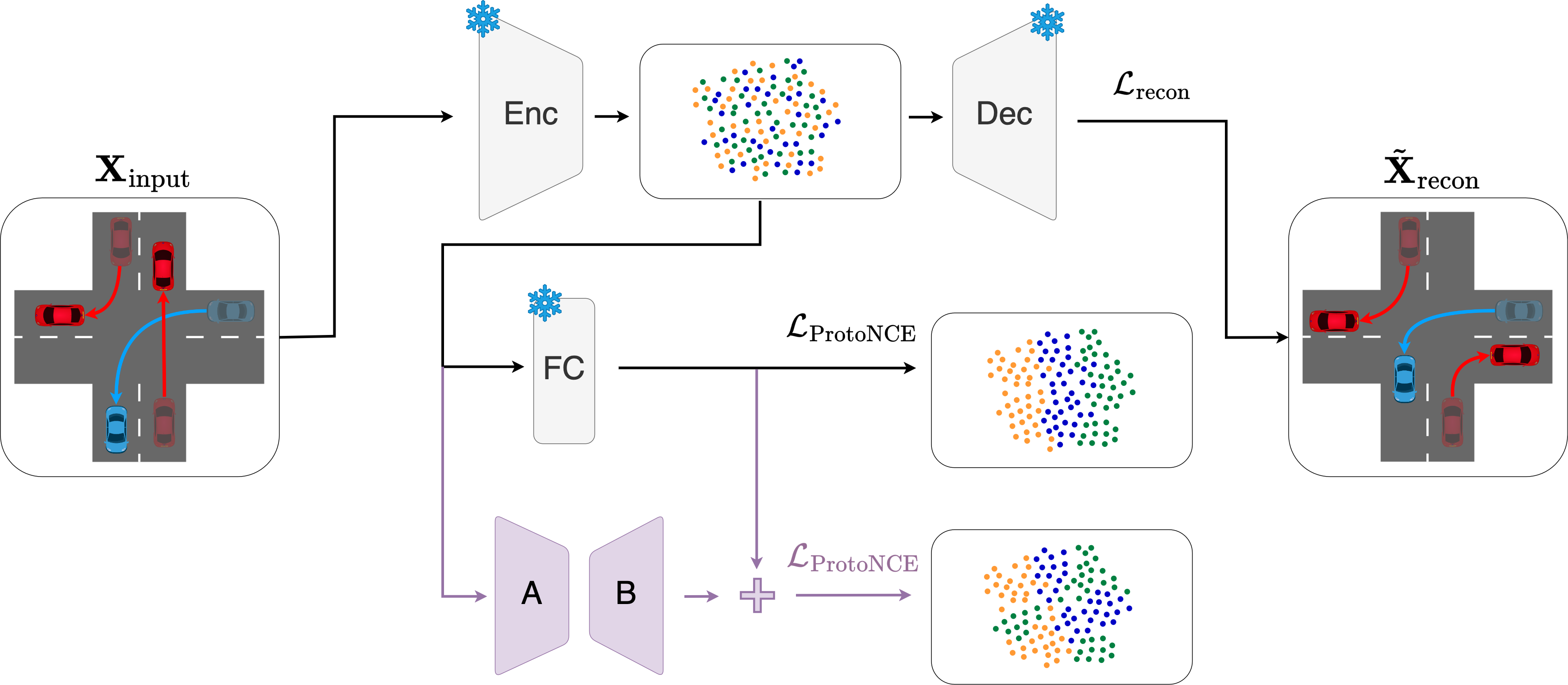}
    \vspace{0.1mm}
    \centering
    \caption{Embedding space training overview, as described in \Cref{sec:approach}. Initial pre-training is performed with reconstruction and contrastive loss objectives. Existing weights are then frozen, and adapters are trained on \texttt{TADS-traj} using contrastive loss alone (shown in \textcolor{lilac}{purple}). Contrastive regularization is supervised via \textcolor{safegreen}{``safe''}, \textcolor{neutralblue}{``neutral''}, and \textcolor{unsafeorange}{``unsafe''} labels.}
    \label{fig:system}
\end{figure*}

Understanding safety-relevant driving behaviors requires an embedding space that captures how agents act under both benign and critical conditions. We aim to learn such a space in a way that supports downstream adversarial scenario-generation, by embedding agent trajectories such that unsafe behaviors are meaningfully structured and distinguishable from safe or neutral ones. This section describes how we construct this safety-aware embedding through four key stages. First, we train a large-capacity, agent-centric model on large-scale trajectory data using a proxy reconstruction objective, producing a general-purpose behavioral encoder. Then, we derive heuristic safety labels and apply contrastive regularization to organize the embedding space and define a structured notion of unsafe behavior. Finally, we refine this space by fine-tuning on a curated set of real-world crash trajectories, adapting the representation to better reflect real safety-critical interactions. This overall process is highlighted in \Cref{fig:system}.

\subsection{Behavior Encoding}\label{ssec:behavior_encoding}

We begin by constructing a large-capacity encoder that produces dense representations of agent behavior from observed trajectories. Our backbone builds on the Motion Transformer (MTR)~\cite{shi2022motion} framework, a state-of-the-art encoder-decoder approach originally designed for multi-agent trajectory prediction. To adapt this architecture for our purposes, we make two key modifications: 1) we replace its agent history conditioning with a full-trajectory conditioning, enabling the encoder to capture complete behavioral signatures; and 2) we remove map conditioning from the input, to support learning from lower-fidelity settings (\idest real-world accident videos) where map information may be unreliable or unavailable. These changes preserve the strong interaction modeling and expressive capacity of MTR while tailoring it to our goal of behavior representation.

Hence, given a scenario $S=(\mathbf{X}, \mathbf{M}, \texttt{ego}, \texttt{adv})$ and a target behavior $X_i \in \mathbf{X}$, we follow the standard MTR encoder structure to compute a context-conditioned behavior embedding $z_i = \text{Enc}(X_i,\ \mathbf{X})$, using full trajectories in $\mathbf{X}$ transformed to the local reference frame of agent $i$ at a fixed timestep. We retain the original MTR decoder architecture, $\text{Dec}(z_i)$, and training setup, using its multi-modal output head to reconstruct the input trajectory from $z_i$, creating $\mathbf{\tilde{X}}$. The framework produces an overall reconstruction loss, $\mathcal{L}_{\text{recon}}$, comprising trajectory regression and mode selection losses.

\subsection{Safety Classification}\label{ssec:safety_classification}

To utilize observed behaviors $\mathbf{X}$ in a safety-informed manner, we first derive large-scale safety labels leveraging the SafeShift~\cite{stoler2024safeshift} scenario characterization framework. For each agent, we compute heuristic safety-relevance scores by aggregating various low-level indicators (\exempli time-to-collision, time headway, trajectory anomaly detection, etc.), applied to both the observed trajectory (``ground-truth''), and a counterfactual extrapolation in which the agent in question proceeds passively (``future extrapolated''; \idest continuing at constant velocity in its current lane). We denote the resulting scores as $\text{GT}_i$ and $\text{FE}_i$, respectively.

Next, to estimate the impact of an agent's \textit{active} behavior on downstream safety, we compute the difference between these two scores. This difference, $d_i = \text{GT}_i - \text{FE}_i$, is then discretized into three classes---``safe'', ``neutral'', and ``unsafe''---according to \Cref{eq:label}, where the thresholding value $\delta$ is selected to ensure roughly even class distributions across the dataset:

\begin{equation}\label{eq:label}
y_i =
\begin{cases}
\text{safe,} & \text{if } d_i < -\delta \\
\text{neutral,} & \text{if } |d_i| \leq \delta \\
\text{unsafe,} & \text{if } d_i > \delta
\end{cases}
\end{equation}

Importantly, although counterfactual alternatives are used to compute these labels, we do not use counterfactual data for training. The representation is trained entirely on unaltered, observed behaviors, preserving the realism of the input trajectories.

\subsection{Contrastive Regularization}\label{ssec:contrastive_regularization}

To organize the learned representations around safety-relevant behavior, we incorporate supervised contrastive learning via the above labels. Architecturally, we project the $z_i \in \mathbf{Z}$ embeddings obtained from $\text{Enc}(X_i, \mathbf{X})$ through an \textit{additional} fully-connected (FC) layer, denoted $\text{FC}_{\text{proj}}$, as in prior work \cite{wang2023fend, makansi2021exposing}. Then, we $\ell_2$-normalize the resulting vector, to obtain $v_i$.
In this way, we can perform regularization on the space spanned by all embeddings $v_i \in \mathbf{V}$ without disrupting the core features in the $\mathbf{Z}$-space which are useful for trajectory reconstruction.

We utilize the standard $\mathcal{L}_{\text{ProtoNCE}}$ objective pioneered in PCL~\cite{li2021prototypical}, and applied in TrACT~\cite{zhang2024tract} and FEND~\cite{wang2023fend}, leveraging both prototype-level and instance-level contrastive losses: $\mathcal{L}_{\text{ProtoNCE}} = \mathcal{L}_{\text{inst}} + \mathcal{L}_{\text{proto}}$.

\noindent That is, for a particular batch of size $B$: %
\begin{equation}\label{eq:inst_cl}
    \mathcal{L_{\text{inst}}} = -\sum_{i=1}^B \frac{1}{N_{\text{po}, i}} \sum_{i^+=1}^{N_{\text{po},i}} \log \frac{\exp(v_i \cdot v_i^+/\tau)}{\sum_{j=1}^B\exp(v_i \cdot v_j/\tau)},
\end{equation} %
\noindent where, for a particular sample $i$, $N_{\text{po},i}$ refers to the number of positive samples (\idest samples with the same class label) within the current batch, $i^+$ indexes one such positive sample, and $\tau$ is a hyperparameter controlling the temperature of the loss. Since all $v_i$ values are normalized, taking their dot product ``$\cdot$'' serves to indicate similarity. The prototypical loss term is defined as follows: %
\begin{equation}\label{eq:proto_cl}
    \mathcal{L_{\text{proto}}} = -\sum_{i=1}^B \log \frac{\exp(v_i \cdot c_i/\phi_i)}{\sum_{j=1}^3\exp(v_i \cdot c_j/\phi_j)},
\end{equation}%
\noindent where $c_i$ and $\phi_i$ denote the prototype centroid and concentration associated with sample $i$’s class, and $c_j$ and $\phi_j$ are those corresponding to each safety class $j \in \{\text{safe},\ \text{neutral},\ \text{unsafe}\}$. Both values are updated once per epoch; centroids $c_j$ are smoothed via momentum with coefficient $\eta$ to promote stability. Centroids are computed over all current embeddings $v_i$ associated with a particular class, while concentration estimates $\phi_j$ are defined as follows, where $N_{\text{class},j}$ denotes the total number of samples associated with class $j$ and $\alpha$ is a non-negative scalar hyperparameter: %
\begin{equation}\label{eq:density}
   \phi_j = \frac{\sum_{i=1}^{N_{\text{class},j}}||v_i - c_j||_2}{N_{\text{class},j}\log(N_{\text{class},j}+\alpha)} 
\end{equation} %
Combined, these two terms have a synergistic effect in organizing the representation space, guiding the space to reflect both fine-grained similarities within safety classes and broader structural separation between them. Our overall training procedure is then to optimize the followinge multi-objective loss: %
\begin{equation}\label{eq:loss}
    \mathcal{L} = \mathcal{L}_{\text{recon}} + \lambda \mathcal{L}_{\text{ProtoNCE}},
\end{equation}
\noindent where $\lambda$ is a tunable hyperparameter that balances reconstruction fidelity against the strength of the safety‐aware contrastive regularization.

\begin{figure*}[t]
    \includegraphics[width=\textwidth]{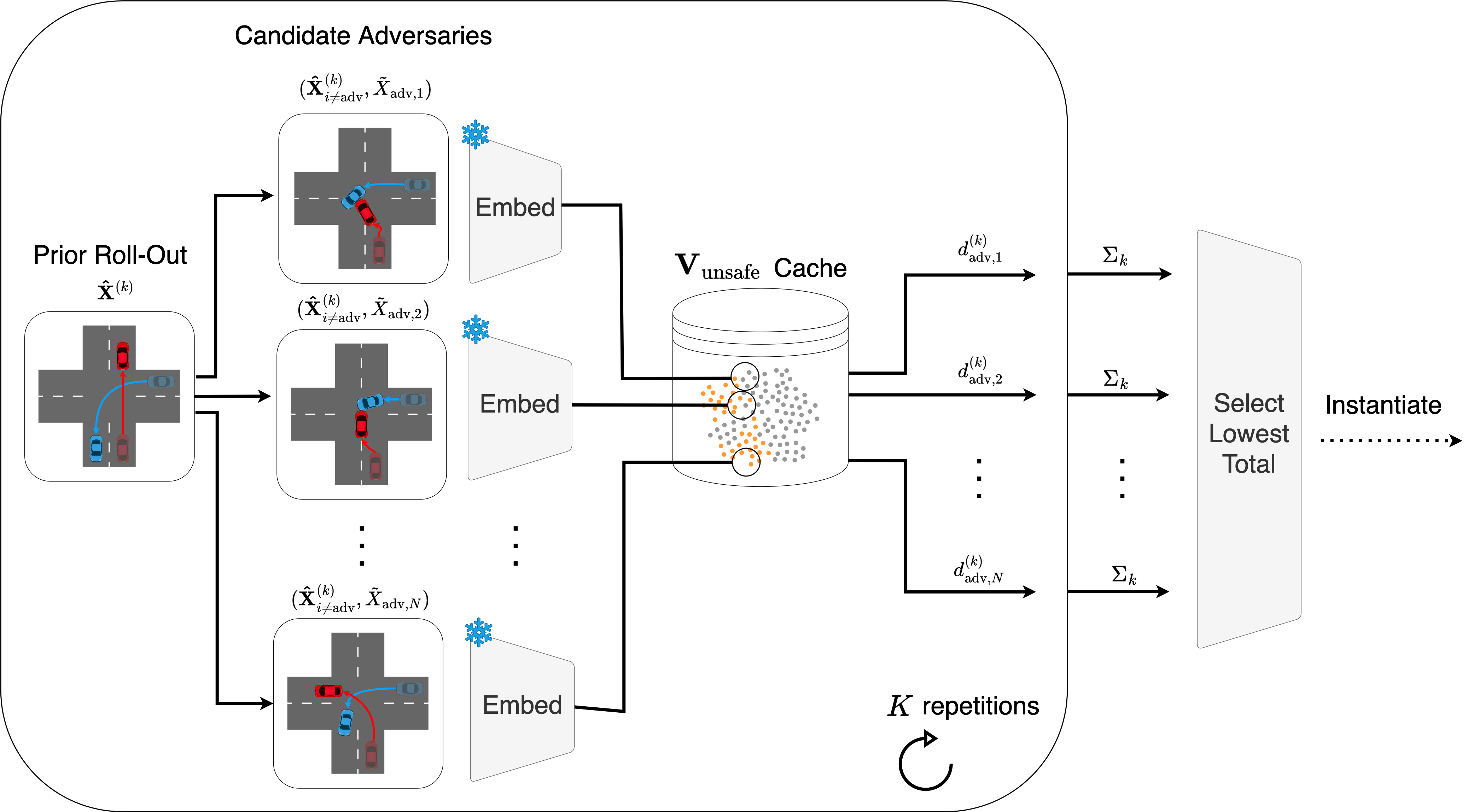}
    \centering
    \vspace{0.1mm}
    \caption{RCG adversarial scenario selection, as described in \Cref{sec:advgen}. Candidate behaviors supplant observed behaviors in $K$ prior roll-outs, then are projected into the $\mathbf{V}$-space and ranked by $k$-NearestNeighbor distance to \textcolor{unsafeorange}{``unsafe''} embeddings.}
    \label{fig:advgen}
\end{figure*}

\subsection{Fine-Tuning Representations}\label{ssec:fine_tuning}

We now aim to calibrate our learned behavior representation by fine-tuning using the small but highly-informative dataset processed in \Cref{sec:dataset}, \texttt{TADS-traj}. Prior work has shown that adapter-based fine-tuning approaches are more robust to overfitting than full fine-tuning, particularly in low-data regimes~\cite{he-etal-2021-effectiveness, lin2024lora}. Indeed, empirically we observe that full fine-tuning leads to rapid memorization and degradation of generalization performance. 
To mitigate this, we employ a LoRA~\cite{hu2022lora} adapter at the encoder bottleneck, inserted in parallel to the contrastive projection head, $\text{FC}_{\text{proj}}$, to act upon $z_i$.

Given $z_i \in \mathbb{R}^{d_1}$ and $\text{FC}_{\text{proj}}(z_i) \in \mathbb{R}^{d_2}$, we follow the original LoRA formulation, introducing a learnable down-projection matrix $A \in \mathbb{R}^{d_1 \times r}$ and up-projection matrix $B \in \mathbb{R}^{r \times d_2}$, where $r$ is the adapter rank. The adapter output is added to the original projection, and the result is $\ell_2$-normalized to yield the final embedding:

\begin{equation}\label{eq:lora}
v_i = \text{normalize} \left( \text{FC}_{\text{proj}}(z_i) + BAz_i \right)
\end{equation}

The up-projection weights $B$ are initialized to zero, so that $v_i$ initially matches the pretrained projection. During fine-tuning, only $A$ and $B$ are updated; all other encoder and decoder parameters remain frozen. Training is performed via the contrastive loss $\mathcal{L}_{\text{ProtoNCE}}$ alone, encouraging additional structure in the representation $\mathbf{V}$-space while preserving the generality of the pre-trained backbone.

%% file: sections/6_advgen.tex
\section{Real-World Crash-Grounded Adversaries} \label{sec:advgen}

To generate realistic yet critical driving scenarios, we propose \textbf{R}eal-world \textbf{C}rash \textbf{G}rounding (RCG) as an adversarial selection mechanism, defined over our behavior embedding space created in \Cref{sec:approach}. Because this embedding space captures high-level semantic structure among agent behaviors, organizing them by safety class while still preserving local variations, we propose a distance measure based on $k$-NearestNeighbors (KNN) to $N_{\text{KNN}}$ known ``unsafe'' embeddings. We then use this KNN-based distance measure to select over \textit{candidate} adversarial behaviors to roll-out against the ego agent, operationalizing the safety-critical perturbation function $\mathcal{P}$ introduced in \Cref{sec:preliminaries}. This overall process is visually represented in \Cref{fig:advgen}.

\noindent \textbf{Behavior Scoring.} Recall that $\mathcal{P}: (S, \{\hat{\mathbf{X}}^{(k)}\}_{k=1}^{K}) \rightarrow \mathcal{B}_{\texttt{adv}}$ maps a base scenario $S$ and up to $K$ historical roll-outs $\{\hat{\mathbf{X}}^{(k)}\}_{k=1}^K$ to a desired adversary behavior $\mathcal{B}_{\texttt{adv}}$, which is then rolled-out in $S$ to construct the next training or evaluation scenario.  We begin by sampling a set of $N_{\text{cand}}$ candidate adversary trajectories from a pre-trained trajectory predictor $\pi_{\text{gen}}$, which models plausible future trajectories conditioned on a fixed history portion of $S$, following the formulation in \cite{zhang2023cat, stoler2024seal}; that is, we collect $\{\tilde{X}_{\text{adv},i}\}_{i=1}^{N_{\text{cand}}} \sim \pi_{\text{gen}}(X_{\texttt{adv}} \mid S)$. 
For each candidate $\tilde{X}_{\text{adv},i}$ and each historical ego trajectory $\hat{X}_{\text{ego}}^{(k)}$, we construct a perturbed scenario $S_i^{(k)}$ by replacing the ego and adversary trajectories in the original scenario’s trajectory set $X$, as in CAT~\cite{zhang2023cat}, and obtain its embedding $v_{\text{adv}, i}^{(k)}$ using the encoder defined in \Cref{ssec:fine_tuning}.

To enable scoring, we precompute and cache the embeddings of all ``unsafe'' base scenarios, denoted $\mathbf{V}_{\text{unsafe}}$. This frozen cache provides a consistent empirical reference for evaluating the semantic plausibility of proposed adversarial behaviors. For each candidate, we then compute the average KNN distance across the $K$ perturbed scenarios:
\begin{equation}\label{eq:knn_dist}
    d_{\text{adv},i} = \frac{1}{K} \sum_{k=1}^{K} \texttt{KNN\_dist}(v_{\text{adv}, i}^{(k)}, \mathbf{V}_{\text{unsafe}})
\end{equation}
\noindent where $\texttt{KNN\_dist}$ denotes the mean Euclidean distance to the embedding's $N_\text{KNN}$ nearest neighbors in $\mathbf{V}_{\text{unsafe}}$.

\noindent \textbf{Behavior Selection.} Since ``unsafe'' in our framework denotes general behaviors that elevate criticality, rather than a scalar notion of risk, emphasizing \textit{local} similarity via KNN produces a more precise and flexible objective than distance to class-level prototypes. To further increase interaction potential, we apply a heuristic collision-closeness score to each candidate, following~\cite{zhang2023cat}, defined as the minimum distance from each historical ego trajectory $\hat{X}_\text{ego}^{(k)}$ to a candidate $\tilde{X}_{\text{adv}, i}$, averaged over the $K$ rollouts. We retain the $N_{\text{int}} < N_{\text{cand}}$ candidates with the lowest average collision-closeness, forming the set $\{\tilde{X}_{\text{adv}, i}\}_{i=1}^{N_{\text{int}}}$, and finally select the one with the minimum $d_{\text{adv},i}$ from this set as $\tilde{X}_{\text{adv}}^*$.

This procedure ensures that the selected adversary behavior is both contextually plausible (in embedding space) and likely to be physically proximate to the ego (in trajectory space, subject to ego reactivity), promoting meaningful and nontrivial interaction. The final behavior functional $\mathcal{B}_{\texttt{adv}}$ is instantiated using the selected trajectory $\tilde{X}_{\texttt{adv}}^*$, and may correspond to open-loop behavior (as in \cite{zhang2023cat}) or a closed-loop reactive policy that utilizes the trajectory as a goal (as in \cite{stoler2024seal}).

%% file: sections/7_experimental_setup.tex
\begin{figure*}[hbtp]
    \centering
    \begin{subfigure}{0.45\textwidth}
        \includegraphics[trim={0 85 0 85},clip, width=\textwidth, page=1]{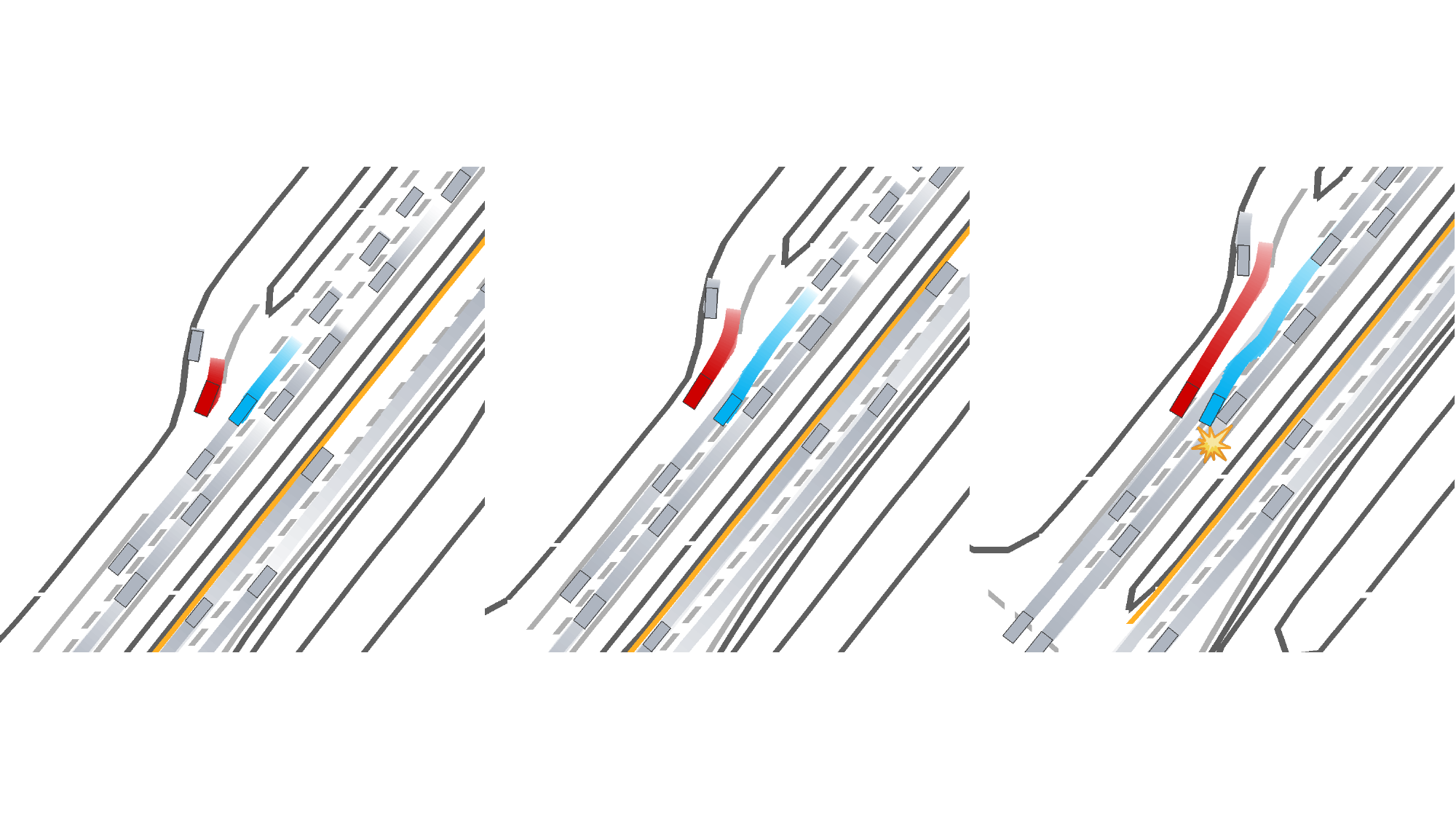}
        \caption{SEAL trained, scenario 1}
    \end{subfigure}
    \hspace*{\fill}
    \begin{subfigure}{0.45\textwidth}
        \includegraphics[trim={0 85 0 85},clip, width=\textwidth, page=2]{figures/ego_viz_qual.pdf}
        \caption{\textbf{SEAL-RCG} trained, scenario 1}
    \end{subfigure}
    \begin{subfigure}{0.45\textwidth}
        \includegraphics[trim={0 85 0 85},clip, width=\textwidth, page=3]{figures/ego_viz_qual.pdf}
        \caption{SEAL trained, scenario 2}
    \end{subfigure}
    \hspace*{\fill}
    \begin{subfigure}{0.45\textwidth}
        \includegraphics[trim={0 85 0 85},clip, width=\textwidth, page=4]{figures/ego_viz_qual.pdf}
        \caption{\textbf{SEAL-RCG} trained, scenario 2}
    \end{subfigure}
\caption{Qualitative examples of closed-loop ego driving in \textbf{unmodified} hard scenarios from WOMD. In each row, the left depicts ego agents trained with SEAL alone; the right shows agents trained using RCG (\Cref{sec:advgen}) instantiated with SEAL. In Scenario 1 (top), our \textcolor[HTML]{00B0F0}{\textbf{ego}} agent avoids a merging \textcolor[HTML]{CD0000}{\textbf{adversary}} while the SEAL-only agent overreacts and crashes into a \textcolor[HTML]{999999}{\textbf{background}} vehicle. In Scenario 2 (bottom), our agent navigates a double left-turn cleanly, while the SEAL-only agent veers too wide and leaves the drivable surface.}
\label{fig:qualitative_ego}
\end{figure*}

\section{Experiment Setup} \label{sec:experimental_setup}

We validate our approach through a sequence of experiments designed to address three research questions:

\begin{itemize}
\item \textbf{RQ1:} Does training the ego agent against RCG adversaries result in more robust closed-loop performance?
\item \textbf{RQ2:} Does our learned behavior embedding meaningfully organize behaviors in a safety-aware manner?
\item \textbf{RQ3:} Does RCG adversary selection lead to more plausible and effective perturbations than baseline methods?
\end{itemize}

Each of the following subsections supports one of these research questions, with RQ1 to \Cref{ssec:rl_impl}, RQ2 corresponding to \Cref{ssec:embedding_impl}, and RQ3 to \Cref{ssec:advgen_impl}. Ablation studies in \Cref{sec:results} further support the validity of these design choices. Although we evaluate all components of the pipeline, our downstream ego training experiments addressing RQ1 provide the most direct and substantial empirical validation of the method’s impact.

\input{tables/main_v2}

\input{tables/aggregate_v1}

\input{tables/ablation_v3}

\subsection{Closed-Loop Ego Training}\label{ssec:rl_impl}

To evaluate the downstream utility of our behavior embedding, we perform closed-loop training of an ego agent in the generated safety-critical scenarios from \Cref{sec:advgen}. This directly assesses whether scenarios generated by our approach, compared to those from prior SOTA baselines (\idest GOOSE~\cite{ransiek2024goose}, CAT~\cite{zhang2023cat}, and SEAL~\cite{stoler2024seal}), provide more effective training stimuli, when tested in both unmodified and adversarially-perturbed scenarios.

We adopt the training and evaluation setup described in SEAL~\cite{stoler2024seal}, using MetaDrive~\cite{li2022metadrive} as the simulator and focusing on the reinforcement learning of an \texttt{ego} agent in non-trivial but non-critical base scenarios from WOMD~\cite{ettinger2021large}. The learning framework, as established in SEAL, builds on ReSkill~\cite{rana2023residual}, a hierarchical reinforcement learning approach that combines offline skill learning with online residual fine-tuning. A continuous skill space is learned from heuristic expert demonstrations, while a high-level policy maps observations (including simulated LiDAR, navigation goals, and ego localization) to latent skill vectors at 1 Hz. A low-level policy, conditioned on the current skill, maps observations at 10 Hz to steering and acceleration commands. During episode roll-out, the skill space and both policies are frozen, while a residual policy is trained online to refine the low-level actions, producing a final control signal to send to the simulator.

Following SEAL, we use 400 base scenarios for training, with 100 held-out \texttt{WOMD-Normal} scenarios and 100 additional safety-relevant base scenarios (\texttt{WOMD-Hard}) for evaluation, performing training and evaluation over four independent seeds. We report performance on both sets of unmodified scenarios, as well as on perturbed versions of the \texttt{WOMD-Normal} scenes generated by each baseline method and our own approach. Evaluation focuses on ego safety metrics, detailed further in \Cref{ssec:rl_results}.

\subsection{Embedding Space Creation Details}\label{ssec:embedding_impl}

When implementing \Cref{sec:approach}, we pre-train our representation space using the WOMD~\cite{ettinger2021large} dataset, leveraging the MTR~\cite{shi2022motion} implementation and training tools provided in UniTraj~\cite{feng2024unitraj}. To ensure compatibility with this pipeline, we convert our \texttt{TADS-traj} dataset into the ScenarioNet~\cite{li2023scenarionet} format, as required by UniTraj. We apply a prototype momentum coefficient of $\eta = 0.8$, contrastive temperature $\tau = 0.05$, and set the concentration parameter $\alpha = 10$, following~\cite{li2021prototypical}. The contrastive loss weight is tuned to $\lambda = 10$.

While LoRA is typically used in low-rank regimes, our architecture's relatively small hidden sizes (\idest $d_1=256$ and $d_2=16$) permits full-rank adaptation ($r=16$). We find this performs better than lower values for $r$; in our case, LoRA primarily mitigates overfitting from full fine-tuning, consistent with findings that LoRA may under-perform in low-dimensional settings~\cite{lialin204relora}.

To assess the learned embedding space, we evaluate its qualitative structure and quantitative properties, comparing multiple configurations and ablations of the training pipeline, as described in \Cref{ssec:embedding_results}.

\subsection{Adversarial Generation Details} \label{ssec:advgen_impl}

We generate adversarial candidate trajectories, as described in \Cref{sec:advgen}, using a pre-trained DenseTNT~\cite{gu2021densetnt} model as the generative policy $\pi_{\text{gen}}$, following prior work~\cite{zhang2023cat, stoler2024seal}. For each adversary agent, we sample $N_\text{cand} = 32$ candidate trajectories and retain the top $N_\text{int} = 6$ highly interactive behaviors.

To instantiate adversarial behaviors, we integrate our grounded objective into both CAT~\cite{zhang2023cat} and SEAL~\cite{stoler2024safeshift} pipelines by directly replacing their respective trajectory scoring objectives. Candidate selection is performed using the distance-based criterion in \Cref{eq:knn_dist}, with $N_\text{KNN}=8$ neighbors for SEAL and $N_\text{KNN}=15$ for CAT. The larger neighborhood for CAT reduces over-reliance on local structure during candidate selection, since it cannot adapt during roll-out; in contrast, SEAL tolerates smaller neighborhoods due to its ability to adjust behavior online.

We perturb scenarios in an iterative manner, up to a maximum number of previous roll-outs $K=5$, following prior work~\cite{zhang2023cat, stoler2024seal}. We evaluate the quality of the generated adversarial trajectories $\hat{X}_\text{adv}^{(K)}$ based on their interaction characteristics, criticality with respect to the ego agent (\idest induced lack of safety), as well as Wasserstein distance (WD)–based realism, following the protocol from SEAL~\cite{stoler2024seal} and detailed in \Cref{ssec:advgen_results}.

\begin{figure*}[t]
\centering
\begin{subfigure}{0.23\textwidth}
    \includegraphics[width=\textwidth, trim=20 20 20 20, clip]{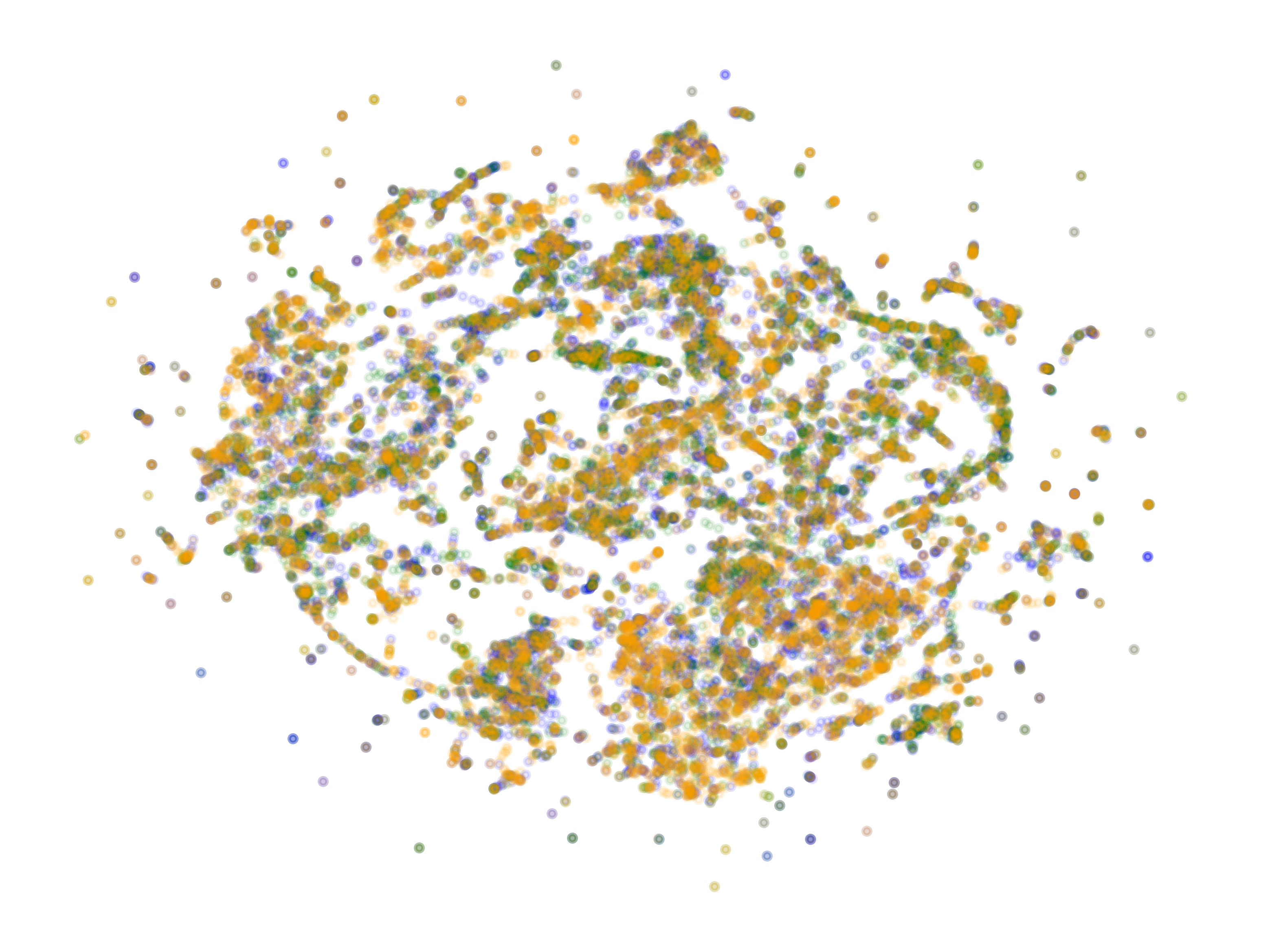}
    \centering
    \caption{No CL}
    \label{sfig:no-pcl}
\end{subfigure}
\hfill
\begin{subfigure}{0.23\textwidth}
    \includegraphics[width=\textwidth, trim=20 20 20 20, clip]{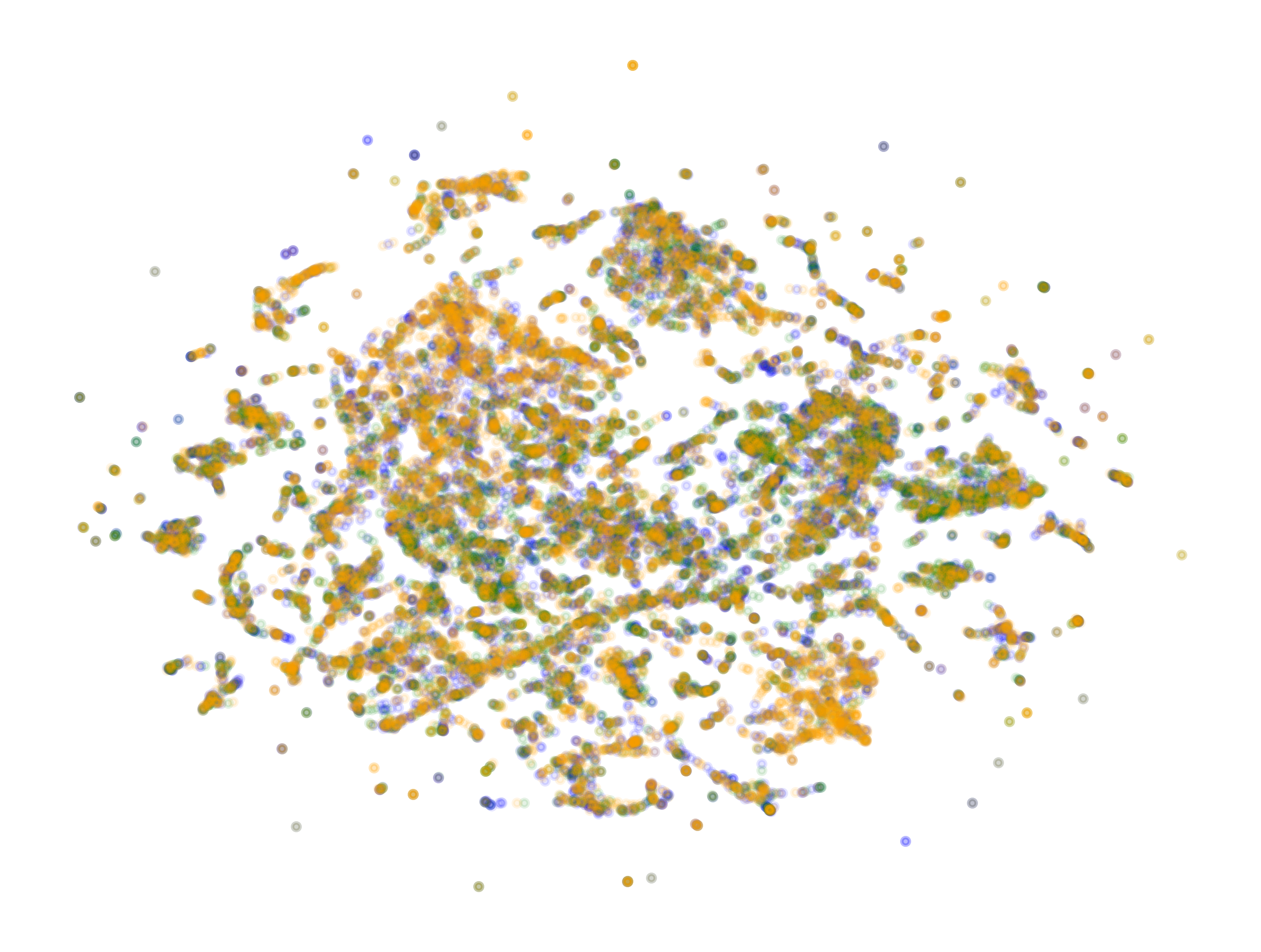}
    \centering
    \caption{\texttt{TADS-traj} FT}
    \label{sfig:no-pcl-tads-ft}
\end{subfigure}
\hfill
\begin{subfigure}{0.23\textwidth}
    \includegraphics[width=\textwidth, trim=20 20 20 20, clip]{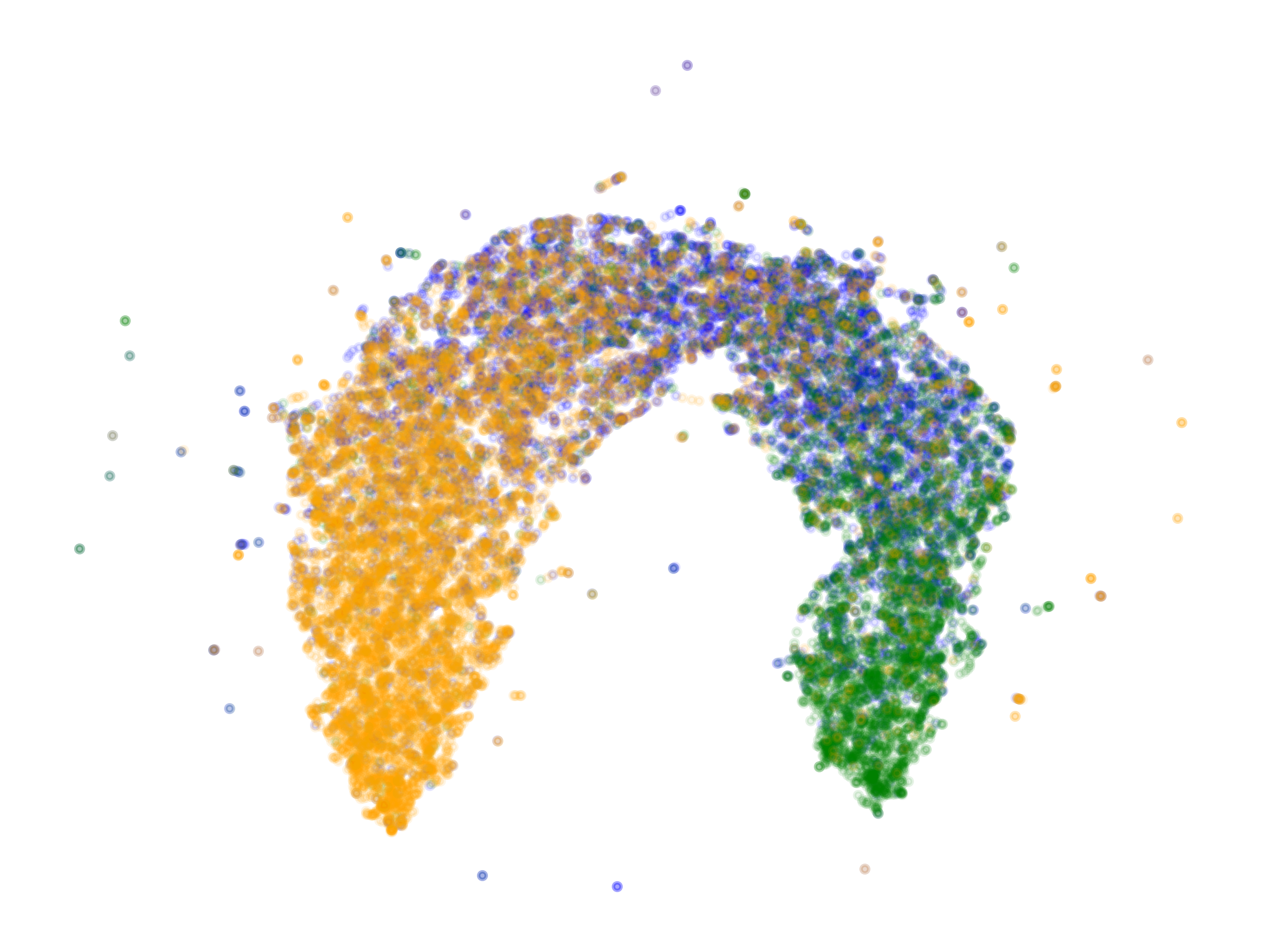}
    \centering
    \caption{CL Pre}
    \label{sfig:womd-cl}
\end{subfigure}
\hfill
\begin{subfigure}{0.23\textwidth}
    \includegraphics[width=\textwidth, trim=20 20 20 20, clip]{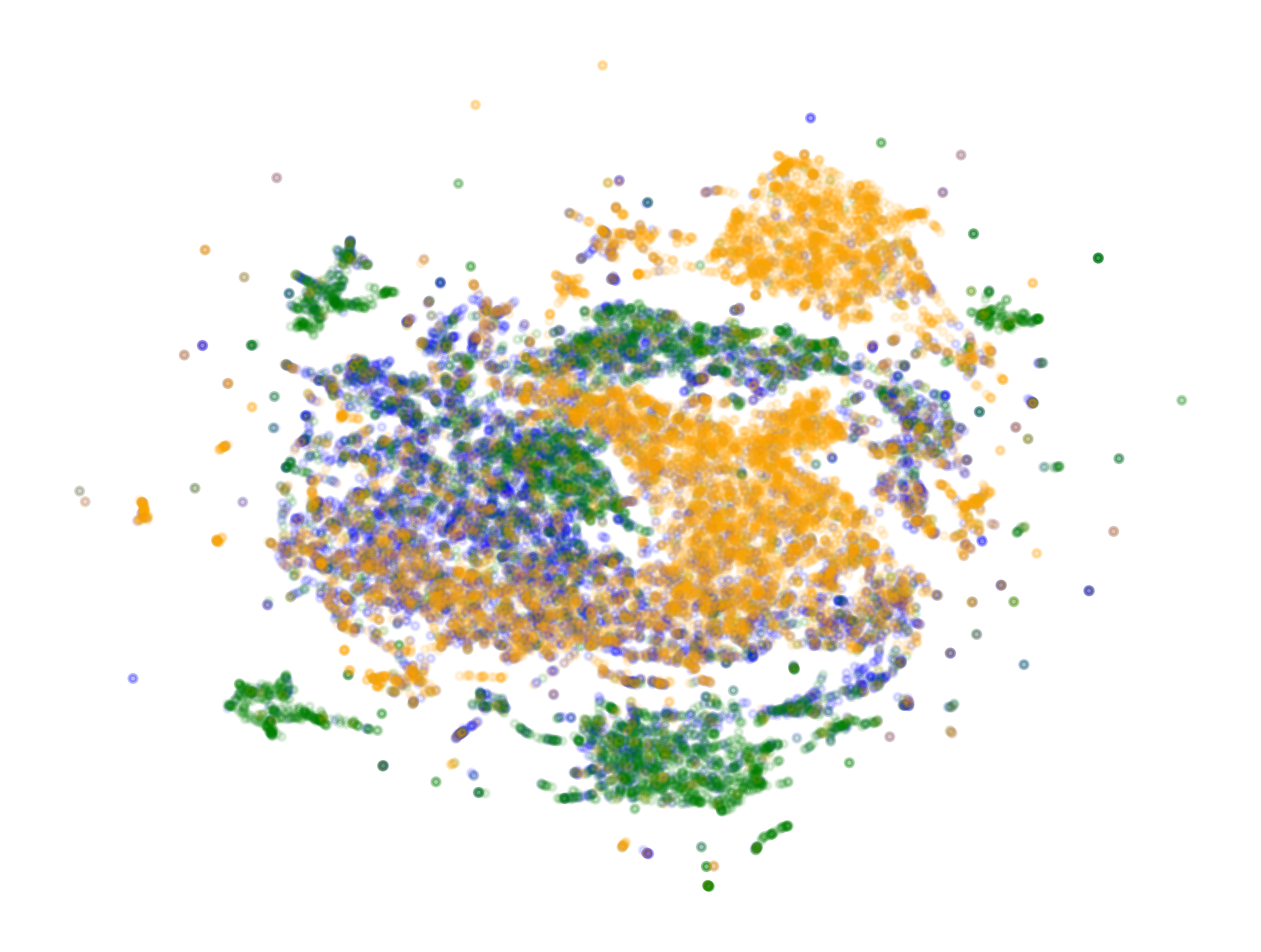}
    \centering
    \caption{CL Pre + \texttt{TADS-traj} FT}
    \label{sfig:tads-ft}
\end{subfigure}
\caption{UMAP~\cite{mcinnes2018umap} projections of the learned representation space, progressing through \Cref{sec:approach}. Each point is an agent behavior, colored by safety-label \textcolor{safegreen}{``safe''}, \textcolor{neutralblue}{``neutral''}, or \textcolor{unsafeorange}{``unsafe''}. Contrastive pre-training with large-scale data (CL, \cref{ssec:contrastive_regularization} and fine-tuning with focused crash data in \texttt{TADS-traj} (FT, \cref{ssec:fine_tuning}) reveal increasingly structured clusters and sub-clusters, aligned with safety semantics.} 
\label{fig:umap_comparison}
\end{figure*}

\input{tables/stats_v1.tex}

%% file: tables/main_v2.tex
\begin{table*}[hbtp]
\centering
\caption{Success rate (\%) of ego agents trained under different scenario generation pipelines, across seven scenario evaluation types. ``Normal'' and ``Hard'' refer to unmodified base scenarios from WOMD~\cite{ettinger2021large}. Columns 3--7 correspond to adversarially-perturbed variants of ``Normal'', each generated by the corresponding method. Each cell reports the mean and standard deviation over four training seeds; higher is better.}

\label{tab:results_main}
\resizebox{0.85\textwidth}{!}{%
\begin{tabular}{llllllll}
\toprule
\textbf{Training Pipeline} & \multicolumn{7}{c}{\textbf{Evaluation Setting}} \\
\cmidrule(lr){2-8}
& Normal & Hard & GOOSE & CAT & SEAL & CAT-RCG & SEAL-RCG \\
\midrule
\textit{None} (Replay)       & 100.0 (0.0) & 97.0 (0.0)  & 59.0 (0.0)  & 18.0 (0.0)  & 32.0 (0.0)  & 49.0 (0.0) & 47.0 (0.0)  \\
\midrule
No Adv          & 49.8 (3.7)  & 29.0 (4.9)  & 41.0 (3.5)  & 31.5 (1.8)  & 31.2 (3.9)  & 35.8 (3.0) & 34.8 (4.7)  \\
GOOSE           & 43.5 (8.1)  & 23.0 (8.2)  & 36.3 (6.2)  & 25.7 (7.0)  & 26.2 (7.9)  & 31.0 (7.2) & 32.8 (7.8)  \\
CAT             & 50.5 (3.6)  & 29.5 (10.7) & 36.8 (5.5)  & 33.0 (6.3)  & 32.0 (3.0)  & 39.2 (6.1) & 36.2 (4.8)  \\
CAT-RCG      & \textbf{52.2 (6.4)}  & \textbf{33.2 (4.5)}  & \textbf{41.2 (2.0)}  & \textbf{38.2 (5.4)}  & \textbf{35.8 (2.9)}  & \textbf{41.8 (2.9)}        &  \textbf{40.0 (6.6)}        \\
\midrule
SEAL     & 54.5 (6.3)  & 36.5 (10.2) & 39.2 (6.9)  & 34.0 (3.2)  & 35.8 (5.1)  & 37.5 (5.2) & 41.0 (4.3)  \\
SEAL-RCG     & \textbf{55.5 (4.0)} & \textbf{36.5 (3.3)} & \textbf{46.0 (2.4)} & \textbf{38.0 (2.7)} & \textbf{38.2 (6.2)} & \textbf{43.8 (2.2)} & \textbf{41.8 (4.4)} \\
\bottomrule
\end{tabular}
}
\end{table*}

%% file: tables/aggregate_v1.tex
\begin{table}[t]
\centering
\caption{Average success rate (\%), crash rate (\%), and out-of-road (OoR) rate (\%) of ego agents over all evaluation settings reported in \Cref{tab:results_main}.}
\label{tab:results_aggregate}
\resizebox{0.45\textwidth}{!}{%
\begin{tabular}{lccc}
\toprule
\textbf{Training Pipeline} & \textbf{Success~$(\uparrow)$} & \textbf{Crash~$(\downarrow)$} & \textbf{OoR~$(\downarrow)$} \\
\midrule
\textit{None} (Replay)     & 57.4 & 42.3 & 00.3 \\
\midrule
No Adv        & 36.1 & 39.0 & 24.9 \\
GOOSE         & 31.2 & 37.9 & 30.9 \\
CAT           & 36.8 & \textbf{28.5} & 34.7 \\
CAT-RCG    & \textbf{40.4} & 28.6 & \textbf{31.0} \\
\midrule
SEAL   & 39.8 & 31.3 & 28.9 \\
SEAL-RCG   & \textbf{42.8} & \textbf{30.7} & \textbf{26.5} \\
\bottomrule
\end{tabular}
}
\end{table}

%% file: tables/ablation_v3.tex
\begin{table}[t]
\centering
\caption{Average performance (\%) of ego agents across ablations. ``CL Pre'' and ``TADS FT'' are contrastive pre-training (\Cref{ssec:contrastive_regularization}) and fine-tuning (\Cref{ssec:fine_tuning}); ``Dist.'' is the distance measure used in \Cref{eq:knn_dist} in \Cref{sec:advgen}. Selected adversary behaviors are instantiated with SEAL.}
\label{tab:results_ablation}
\resizebox{0.48\textwidth}{!}{%
\begin{tabular}{ccc|ccc}
\toprule
\textbf{CL Pre} & \textbf{TADS FT} & \textbf{Dist.} & \textbf{Success~$(\uparrow)$} & \textbf{Crash~$(\downarrow)$} & \textbf{OoR~$(\downarrow)$} \\
\midrule
-- & -- & KNN     & 33.9 & 32.1 & 34.1 \\
\checkmark & -- & KNN  & 38.4 & 30.7 & 30.9 \\
-- & \checkmark & KNN  & 39.2 & \textbf{30.6} & 30.2 \\
\checkmark & \checkmark & Proto  & 41.6 & 31.7 & 26.7 \\
\midrule
\checkmark & \checkmark & KNN    & \textbf{42.8} & 30.7 & \textbf{26.5} \\
\bottomrule
\end{tabular}
}
\end{table}

%% file: tables/stats_v1.tex
\begin{table}[t]
\centering
\caption{Quantitative analysis of embedding space structure. Clustering metrics (Silhouette score and Davies-Bouldin index) are computed over per-class KMeans sub-clustering (averaged over $N_{km} \in \{3, 4, 5, 6\}$), while linear probes assess classification accuracy. Results are averaged over three seeds.}
\resizebox{0.49\textwidth}{!}{
\begin{tabular}{lccc}
\toprule
\textbf{Method} & \textbf{Silhouette} $\uparrow$ & \textbf{Davies-Bouldin} $\downarrow$ & \textbf{Linear Probe Acc.} $\uparrow$ \\
\midrule
No CL        & 0.177 & 1.768 & 44.1\% \\
\texttt{TADS-traj} FT        & 0.171 & 1.734 & 43.6\% \\
CL Pre      & 0.189 & 1.624 & 46.5\% \\
CL Pre + \texttt{TADS-traj} FT      & \textbf{0.198} & \textbf{1.278} & \textbf{46.8}\% \\
\bottomrule
\end{tabular}}

\label{tab:embedding_analysis}
\end{table}

%% file: sections/8_results.tex
\section{Results} \label{sec:results}

\subsection{Closed-Loop Training Results}\label{ssec:rl_results}

\Cref{tab:results_main} shows our main quantitative results, highlighting overall task success rate improvements with RCG. Both SEAL-RCG and CAT-RCG substantially outperform their respective baselines, with an average success rate gain of $9.2\%$, confirming \textbf{RQ1}. Performance is strictly equal or better across all seven evaluation settings, and standard deviations are generally smaller, indicating more consistent behavior. For completeness, RCG-enhanced pipelines also outperform agents trained with GOOSE~\cite{ransiek2024goose}-generated scenarios, as well as those trained without perturbations (``No Adv'').

Qualitatively, \Cref{fig:qualitative_ego} illustrates how ego agents trained with RCG adversaries outperform those trained without. In both scenarios, the baseline SEAL-trained policy overreacts to the adversary---swerving into background vehicles or leaving the drivable area---whereas the SEAL-RCG trained policy successfully navigates the challenging scenario. \Cref{tab:results_aggregate} provides a deeper breakdown by failure type; while RCG-based agents achieve similar crash rates to their respective baselines, they show substantial improvements in out-of-road rates, highlighting that prior approaches struggle to balance both failure modes.

To understand how various components of RCG contribute to final performance, we provide an ablation study in \Cref{tab:results_ablation}, breaking down the training pipeline along components introduced in \Cref{sec:approach} and \Cref{sec:advgen}. First, we ablate the contrastive pre-training described in \Cref{ssec:contrastive_regularization}, training the embedding model with either $\mathcal{L}_\text{recon}$ alone or the full loss from \Cref{eq:loss}. We then ablate the \texttt{TADS-traj} contrastive fine-tuning phase described in \Cref{ssec:fine_tuning}, either omitting or including it. Finally, we ablate the distance measure used for candidate scoring against the embedding cache, as formalized in \Cref{eq:knn_dist}, by replacing the $k$-NearestNeighbor function with distance to the ``unsafe'' prototype center. As shown in the bottom row, all components are required for peak performance.

\begin{figure*}[hbtp]
    \centering
    \begin{subfigure}{0.45\textwidth}
        \includegraphics[trim={0 85 0 85},clip, width=\textwidth, page=1]{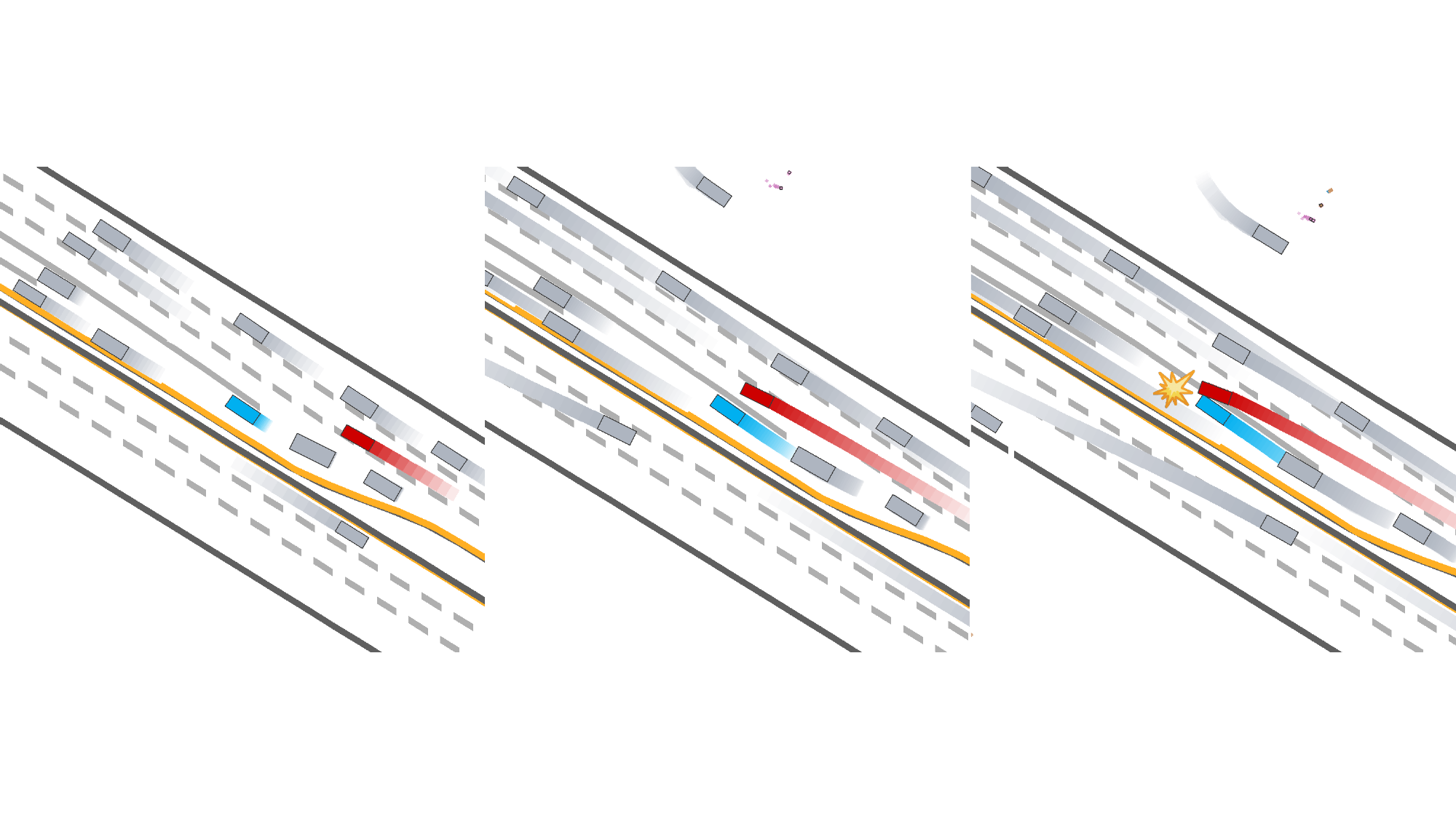}
        \caption{SEAL perturbation, scenario 1}
    \end{subfigure}
    \hspace*{\fill}
    \begin{subfigure}{0.45\textwidth}
        \includegraphics[trim={0 85 0 85},clip, width=\textwidth, page=2]{figures/viz_qual.pdf}
        \caption{\textbf{SEAL-RCG} perturbation, scenario 1}
    \end{subfigure}
    \begin{subfigure}{0.45\textwidth}
        \includegraphics[trim={0 85 0 85},clip, width=\textwidth, page=3]{figures/viz_qual.pdf}
        \caption{SEAL perturbation, scenario 2}
    \end{subfigure}
    \hspace*{\fill}
    \begin{subfigure}{0.45\textwidth}
        \includegraphics[trim={0 85 0 85},clip, width=\textwidth, page=4]{figures/viz_qual.pdf}
        \caption{\textbf{SEAL-RCG} perturbation, scenario 2}
    \end{subfigure}
\caption{Qualitative examples of adversarial \textbf{perturbation} against an ego replay policy. In each row, the left shows perturbations from SEAL alone, and the right shows SEAL-RCG, applied to the same base scenario.
In Scenario 1 (top), the \textcolor[HTML]{CD0000}{\textbf{adversary}} merges directly into the \textcolor[HTML]{00B0F0}{\textbf{ego}} under SEAL, whereas our method produces a near-miss cut-in.
In Scenario 2 (bottom), SEAL causes the ego to t-bone an adversary that drives unrealistically across the lane toward the road edge, whereas our method produces a more subtle, glancing collision as both agents turn into the same lane.}
\label{fig:qualitative_advgen}
\end{figure*}

\begin{figure*}[hbtp]
    \centering
    \begin{subfigure}[t]{0.45\textwidth}
        \centering
        \includegraphics[width=\textwidth]{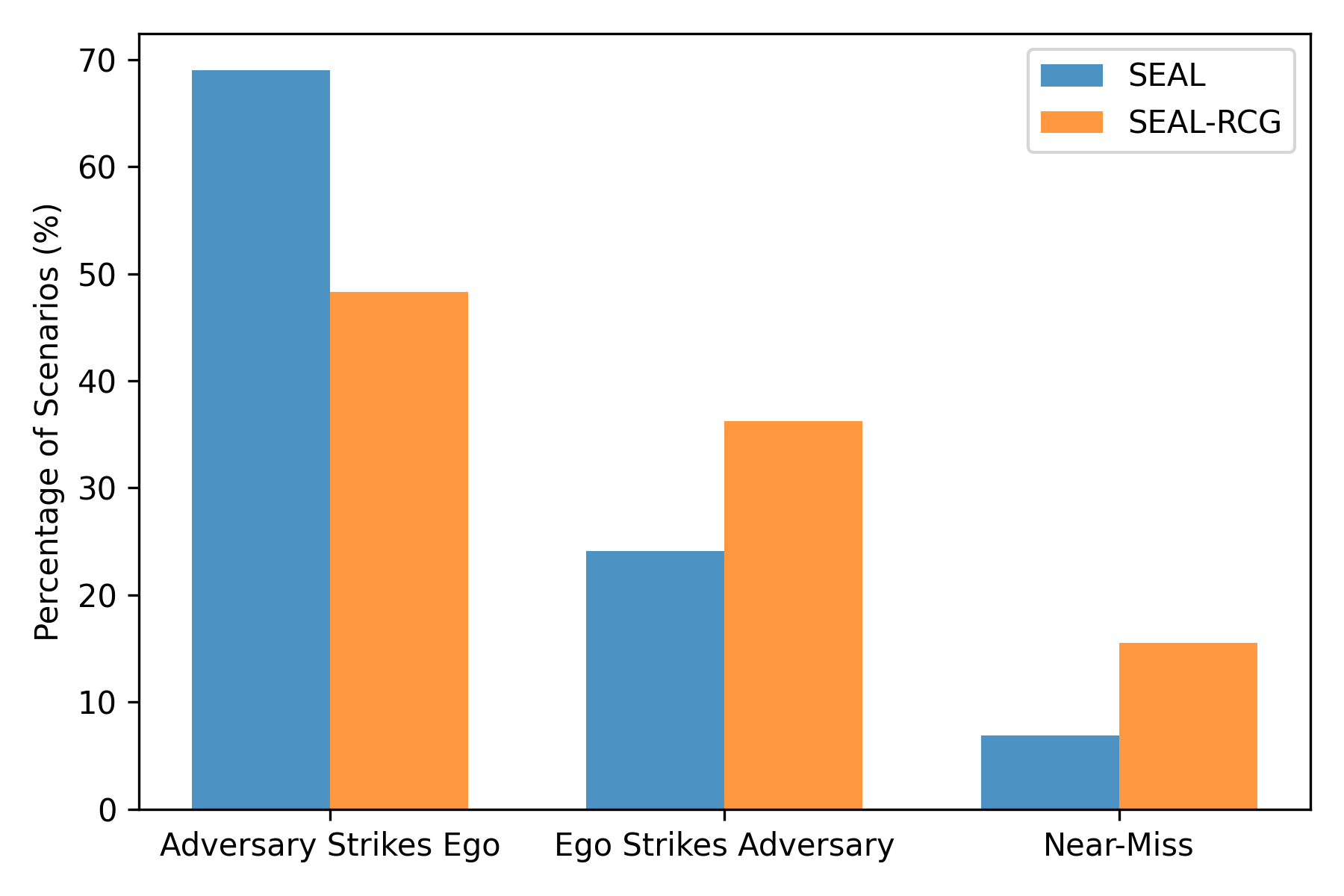}
        \caption{Collision Causality}
        \label{sfig:collision_causality}
    \end{subfigure}
    \hspace*{\fill}
    \begin{subfigure}[t]{0.45\textwidth}
        \centering
        \includegraphics[width=\textwidth]{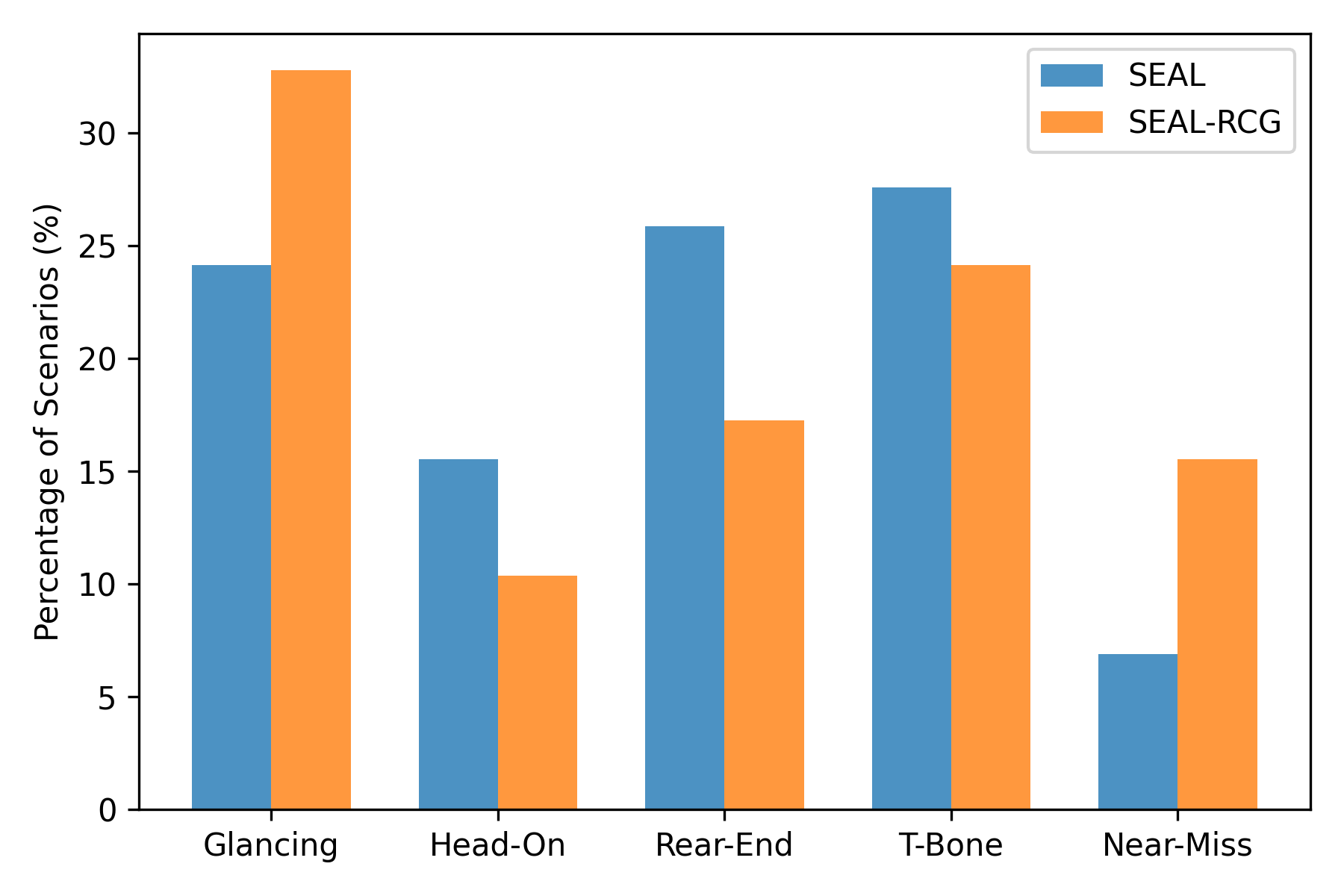}
        \caption{Collision Geometry}
        \label{sfig:collision_geometry}
    \end{subfigure}
    \caption{
    Taxonomizations of generated adversarial interactions across matched base scenarios, against an ego replay policy. }
    \label{fig:scenario_types}
\end{figure*}

\subsection{Embedding Space Analysis}\label{ssec:embedding_results}

We analyze the learned behavior embedding space by evaluating both its qualitative structure and quantitative properties. All evaluations focus on the contrastive feature embeddings $v_i$, derived from the training pipeline described in \Cref{sec:approach} and projected from WOMD input scenarios.

\Cref{fig:umap_comparison} illustrates how the representation evolves across training stages. Without contrastive learning (\Cref{sfig:no-pcl}), the space lacks any clear organization with respect to safety. Applying prototypical contrastive regularization (\Cref{sfig:womd-cl}) introduces coarse semantic structure aligned with safety labels, providing an initial organization where distance reflects behavioral risk. This is essential for downstream use, where candidate behaviors are scored by proximity to known unsafe examples. Fine-tuning on the \texttt{TADS-traj} dataset (\Cref{sfig:tads-ft}) further refines this space, introducing sub-cluster structure that captures more \textit{granular} distinctions within each safety class. This refinement arises because the critical trajectories from TADS introduce higher-severity interactions and contextually unsafe behavior, expanding the variation within each class in ways that the contrastive objective can retain. Conversely, fine-tuning without the initial PCL stage (\Cref{sfig:no-pcl-tads-ft}) fails to meaningfully reshape the space, supporting the claim that our full pipeline is necessary.

Quantitatively, we assess the structure of the embedding space using unsupervised clustering quality and supervised probe informativity. For clustering, we perform KMeans (with $N_{km}$ clusters) within each safety class and report the average Silhouette score and Davies-Bouldin index on these sub-clusters, two standard metrics for capturing intra-class coherence and inter-class separation~\cite{chua2024learning, halkidi2001clustering}. For informativity, we train linear probes to predict safety labels on the same held-out validation set used in \Cref{sec:approach}. Results are reported in \Cref{tab:embedding_analysis}; a higher Silhouette score and probe accuracy is better, while a lower Davies-Bouldin index is better. All results are averaged over three seeds, with clustering metrics further averaged over $N_{km} \in \{3, 4, 5, 6\}$. Overall, our full method improves across all metrics, confirming \textbf{RQ2} by demonstrating the stronger intrinsic capabilities of the learned representation.

\subsection{Generated Scenario Results}\label{ssec:advgen_results}

We highlight representative qualitative examples of adversarial scenario generation in \Cref{fig:qualitative_advgen}, comparing SEAL alone to SEAL-RCG on matched base scenarios. Our method induces more nuanced behaviors, including near-misses and subtle collisions, in contrast to the overtly aggressive and often implausible maneuvers produced by SEAL. To compare broader behavioral characteristics, we analyze perturbation structure across a shared subset of base scenarios where both approaches produce critical or near-critical outcomes, as shown in \Cref{fig:scenario_types}. 
Our approach increases the rate of near-misses and ego-initiated collisions, suggesting that RCG adversaries induce high-risk interactions that are more contingent on ego behavior, rather than forcing direct collisions. Geometrically, our method produces more glancing impacts and fewer t-bone or in-line collisions, better aligning with adversary maneuvers that \textit{reduce} impact severity, as observed in real-world crash data such as \texttt{TADS-traj}.

Quantitatively, as shown in \Cref{tab:gen_quality}, baseline approaches like CAT and SEAL produce more extreme scenarios than RCG-based methods, as the former directly optimize for criticality. Still, as shown in \Cref{ssec:rl_results}, training ego policies on these highly critical examples does not necessarily lead to greater downstream ego performance. We additionally evaluate distributional realism via Wasserstein distances on yaw and acceleration, following SEAL~\cite{stoler2024seal}. Unlike SEAL, we do not penalize adversary out-of-road behavior, as such maneuvers are common in real-world crashes (\exempli in \texttt{TADS-traj}) and do not necessarily indicate implausibility in safety-critical settings. These distances, computed against ground-truth $X_\text{adv}$ trajectories, serve as a rough proxy for low-level behavioral realism. Under this measure, SEAL-RCG achieves the strongest alignment with real-world distributions, with CAT-RCG also showing modest gains. While these metrics capture distributional similarity, they overlook key aspects of interaction \textit{quality}, further motivating the causal and geometric analyses above. Thus, taken together, these qualitative and quantitative results support \textbf{RQ3}.

\input{tables/advgen_v2}

%% file: tables/advgen_v2.tex
\begin{table}[t]
\centering
\caption{
Scenario generation criticality and realism results; \textbf{lower} ego success is better. Wasserstein Distance (WD) on yaw and acceleration reflects alignment with real-world behavior. Averages are computed over all tested ego agents. 
}
\label{tab:gen_quality}
\resizebox{0.5\textwidth}{!}{\begin{tabular}{lllcc}
\toprule
Eval. Setting & Ego Success ($\downarrow$) & Yaw WD ($\downarrow$) & Acc WD ($\downarrow$) \\
\midrule
Normal        & 55.5\%  & 0.134 & 0.032 \\
Hard          & 37.1\%  & 0.122 & 0.050 \\
\midrule
GOOSE         & 42.1\%  & 0.138 & 0.617 \\
CAT           & \textbf{32.1\%}  & 0.136 & 0.291 \\
CAT-RCG    & 39.2\%  & 0.135 & 0.281 \\
SEAL          & 33.2\%  & 0.135 & 0.160 \\
SEAL-RCG   & 39.1\%  & \textbf{0.135} & \textbf{0.117} \\
\bottomrule
\end{tabular}
}
\end{table}

%% file: sections/9_conclusion.tex
\section{Conclusion} \label{sec:conclusion}

Scenario generation is essential for developing and validating autonomous driving systems, but existing approaches often produce interactions that are overly simplistic or behaviorally unrealistic. We introduced Real-world Crash Grounding (RCG), a framework for generating safety-critical scenarios by guiding adversarial perturbations using a behavior embedding grounded in real-world crash data. We integrated RCG into two prior generation pipelines, CAT and SEAL, replacing their handcrafted scoring mechanisms with our embedding-based selection criterion. Ego agents trained against RCG-perturbed scenarios achieved consistently higher success rates, with an average improvement of 9.2\% across seven evaluation settings. Further analysis showed that the resulting scenarios elicited more causally and geometrically nuanced interactions, better reflecting real-world failure modes.

While our approach improves scenario quality and ego robustness, further enhancements remain possible. Scaling to additional crash video sources and leveraging advances in perception models could improve the quality and diversity of approximate trajectory annotations. Future work may also explore ego-side applications of the representation, such as using distance to ``safe'' embeddings to guide ego maneuver selection in closed-loop settings.